\documentclass[11pt,onecolumn]{style}
\pdfmapfile{+assets/fonts/nunito/map/Nunito.map}
\usepackage{amsmath,amssymb,amsthm}
\usepackage{xspace,enumitem,tabularx,array,makecell}
\usepackage{placeins}
\usepackage{wrapfig}
\usepackage{algorithm,algorithmic,longtable,lscape}
\let\ufolandscape\landscape
\let\endufolandscape\endlandscape
\renewenvironment{landscape}{\ufolandscape\pdfpageattr{/Rotate 90}}%
  {\endufolandscape\pdfpageattr{/Rotate 0}}
\usepackage{tikz}
\usetikzlibrary{arrows.meta,positioning}
\newcommand{\method}{UniVEP\xspace}
\newcommand{\E}{\mathbb E}
\newcommand{\R}{\mathbb R}
\newcommand{\norm}[1]{\left\lVert #1\right\rVert}

\newcommand{\papertablestyle}{\fontsize{9}{10.5}\selectfont\renewcommand{\arraystretch}{1.05}}
\newcommand{\rankfirst}[1]{\cellcolor{hiveBestBg}\textcolor{hiveBestText}{\textbf{#1}}}

\newsavebox{\teasercaptionbox}
\newlength{\teaserimageheight}
\newsavebox{\resultspagebox}
\newsavebox{\ufotablebox}
\newenvironment{fitwidth}{\begin{lrbox}{\ufotablebox}}{%
  \end{lrbox}%
  \ifdim\wd\ufotablebox>\linewidth
    \resizebox{\linewidth}{!}{\usebox{\ufotablebox}}%
  \else\usebox{\ufotablebox}\fi}
\renewcommand{\finding}[2]{%
  \par\noindent%
  \refstepcounter{finding}\noindent%
  {\color{#1}\bfseries\interlinepenalty=10000 Finding~\thefinding: #2\par}%
  \nobreak\csname @afterheading\endcsname}
\DeclareMathSizes{7.1}{7.1}{5}{5}
\hypersetup{pdftitle={Unifying Velocity and Endpoint Prediction},
  pdfauthor={Author information to be added}}
\title{UniVEP: Unifying Velocity and Endpoint Prediction for Robot Manipulation}
\author{%
\begin{HIVEauthorblock}
\HIVEauth{Zhengtu Liang}{1}{3,*}\authsep
\HIVEauth{Qiwei Liang}{1}{*,\S}\authsep
\HIVEauthAff{Tao Lin}{2}\authsep
\HIVEauthAff{Shihang Weng}{3}\authsep
\HIVEauthAff{Boyang Cai}{1}\authsep
\HIVEauthAff{Xiangyu Chen}{1}\authbreak
\HIVEauthAff{Yikeng Chen}{1}\authsep
\HIVEauthAff{Jiaming Liang}{1}\authsep
\HIVEauthAff{Bohan Tan}{1}\authsep
\HIVEauthAff{Yu Yan}{1}\authsep
\HIVEauth{Renjing Xu}{1}{\ddag}
\end{HIVEauthorblock}
}

\affiliation{%
\begin{HIVEaffiliationblock}
\affitem{1}{The Hong Kong University of Science and Technology (Guangzhou)} \\[0.15em]
\affitem{2}{Beijing Jiaotong University}\quad
\affitem{3}{Shenzhen University} 
\end{HIVEaffiliationblock}
}

\correspondence{%
  \metamark{*} Equal contribution \HIVEmetasep
  \metamark{\S} Project leader \HIVEmetasep
  \metamark{\ddag} Corresponding author
}

\date{}

\usepackage{needspace}
\AddToHook{cmd/section/before}{\Needspace{6\baselineskip}}
\AddToHook{cmd/subsection/before}{\Needspace{4\baselineskip}}
\newcolumntype{P}{l}
\newcolumntype{R}{>{\raggedleft\arraybackslash}X}
\newcolumntype{C}{>{\centering\arraybackslash}X}
\newcommand{\papertablelook}{\renewcommand{\arraystretch}{1.05}\arrayrulecolor{hiveInk}}
\renewcommand{\papertablestyle}{\fontsize{9}{10.5}\selectfont\papertablelook\setlength{\tabcolsep}{3pt}}
\newcommand{\thdr}{}
\newcommand{\tmethod}{}
\newcommand{\tgroup}{}
\definecolor{hiveGroupBg}{HTML}{F1F5F9}

\newcommand{\arxivfirst}[1]{\cellcolor{hiveBestBg}\textcolor{hiveBestText}{#1}}
\newcommand{\arxivsecond}[1]{\cellcolor{hiveSecondBg}\textcolor{hiveSecondText}{#1}}
\newcommand{\arxivthird}[1]{\cellcolor{hiveThirdBg}\textcolor{hiveThirdText}{#1}}
\newtcolorbox{claim}{enhanced,breakable,colback=hiveSoftBlue,
  boxrule=0pt,frame hidden,borderline west={2.2pt}{0pt}{hiveAccent},
  sharp corners,left=8pt,right=6pt,top=4pt,bottom=4pt,
  before skip=7pt plus 2pt,after skip=6pt plus 2pt,fontupper=\normalsize}
\newcommand{\makeappendixtitle}{%
  \begin{center}{\Large\bfseries Appendix}\end{center}}
\begin{document}
\maketitle

\begin{abstract}
  Flow-matching robot policies usually use one prediction target throughout action generation, even though recovering actions from noise and polishing nearly clean actions bring different challenges. We systematically compare velocity prediction (V), endpoint prediction (X), and velocity-weighted endpoint prediction (X--V) across multiple robot policies, benchmarks, and inference budgets. On StarVLA and LiLa-WAM, X--V works best for few-step generation, while V gains more from iterative refinement. We further uncover their complementary strengths across noise levels: clean actions have concentrated covariance spectra, endpoint predictions stay more stable under high noise, and velocity predictions offer more accurate corrections near clean actions. Inspired by these findings, we introduce \textbf{UniVEP}, which learns both prediction targets together with consistency supervision and switches from endpoint recovery to velocity refinement as noise drops. Each step only needs one network run, and joint training also boosts single-step endpoint generation. Tests on simulated and real-world manipulation show better success rates and efficiency. On RoboTwin, UniVEP hits 61.36\% success with one StarVLA run, beating the best single-target baseline at sixteen runs (59.96\%). With LiLa-WAM, two runs outperform the best sixteen-step single-target baseline and cut inference latency by 77.1\%.
  \href{https://tutuliang904.github.io/UniVEP/}{\textbf{Project page \& Codebase}}
\keywords{$\;\bullet\;$ Robot Manipulation $\;\bullet\;$ Flow Matching $\;\bullet\;$ Prediction Targets}
\end{abstract}

\begin{lrbox}{\teasercaptionbox}
\begin{minipage}{\linewidth}
 \captionsetup{type=figure}
 \caption{\textbf{From studying action prediction to UniVEP.} Our robot action prediction study finds X-pred strong at high noise but weak at low noise, with V-pred showing the opposite. These findings motivate UniVEP, which unifies both prediction modes and achieves strong manipulation performance with only one or two inference steps.}
 \label{fig:teaser}
\end{minipage}
\end{lrbox}
\vspace{5pt}
\nobreak
\setlength{\teaserimageheight}{\dimexpr
  \pagegoal-\pagetotal-\pagedepth-\ht\teasercaptionbox-\dp\teasercaptionbox-8pt\relax}
\nointerlineskip
\vbox{%
  \hbox to \linewidth{\hss
    \includegraphics[width=1.08\linewidth,height=\teaserimageheight,keepaspectratio,
      trim=20bp 19bp 30bp 12bp,clip]{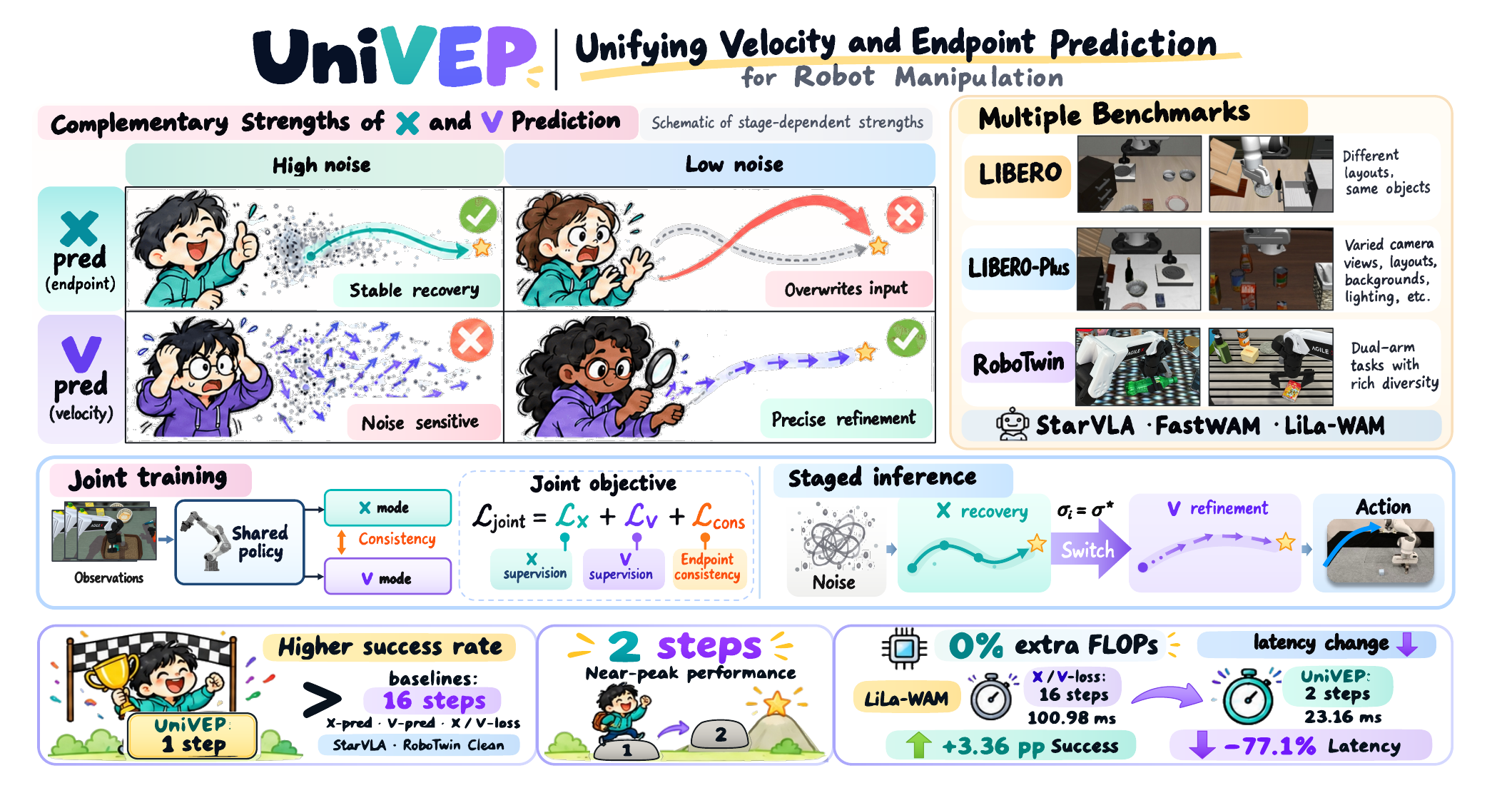}%
    \hss}%
  \nointerlineskip
  \hbox{\usebox{\teasercaptionbox}}%
}
\clearpage
\section{Introduction}
\label{sec:introduction}

Generative robot policies can produce rich, continuous action sequences, but they must do so quickly enough for closed-loop control. Modern vision-language-action (VLA) policies and world-action models (WAMs) commonly use flow matching to transform Gaussian noise into executable action chunks~\citep{black2024pi0,bjorck2025gr00t,yuan2026fastwam,yang2026lila}. Each refinement step requires another network evaluation, creating a trade-off between action quality and control latency. While much effort has gone into reducing sampling steps~\citep{wang2024onedp,prasad2024consistencypolicy,zhang2025flowpolicy}, we revisit a more basic question: \textbf{\emph{what should the action network predict?}}

Most flow-based policies predict the velocity that transports a noisy action toward the data. An alternative is to predict the clean action endpoint directly. Although these two predictions are analytically interchangeable along a linear noise-to-data path~\citep{lipman2023flowmatching,liu2023rectifiedflow,albergo2023interpolants}, they pose different learning problems. Endpoint prediction seeks a structured action, whereas velocity prediction must also account for the injected noise. Moreover, generation itself changes character over time: the model first needs to recover an action from heavy noise, then make increasingly precise corrections as the sample approaches a valid trajectory. There is little reason to expect one prediction target to be equally effective at both stages, from early recovery to final refinement.

We investigate this question by comparing three objectives: velocity prediction (V), direct endpoint prediction (X), and endpoint prediction with a velocity-weighted loss (X--V), as explored in action-manifold learning~\citep{yang2026abot,xiao2026multiviewaml}. Our study covers three policy architectures---StarVLA, FastWAM, and LiLa-WAM~\citep{starvla2026,yuan2026fastwam,yang2026lila}---on LIBERO, LIBERO-Plus, and RoboTwin 2.0~\citep{liu2023libero,fei25libero-plus,chen2025robotwin}. Within each policy, we keep the architecture, training data, and update budget fixed while varying inference from one to sixteen network evaluations. On StarVLA and LiLa-WAM, X--V performs particularly well with one or two evaluations, while V benefits substantially from additional steps. Direct X, in contrast, improves little as the budget grows. The best prediction objective therefore depends not only on the policy, but also on how many refinement steps it can afford.

Why does this ranking change? We examine the structure of the targets and the behavior of their predictions along the noise-to-action path. Clean action chunks have far more concentrated sample covariance spectra than velocity targets or Gaussian noise. At high noise, endpoint predictions remain substantially more stable across different noise initializations. Near a clean action, velocity-based updates produce lower recovery error: they correct the current estimate rather than predicting the entire action again. These results point to a natural division of labor (\cref{fig:teaser}): \textbf{\emph{endpoint prediction for recovering actions from noise, and velocity prediction for refining them near the end of generation.}}

Guided by this finding, we introduce \textbf{UniVEP} (\textbf{Uni}fying \textbf{V}elocity and \textbf{E}ndpoint \textbf{P}rediction). UniVEP trains both prediction modes within a shared action network, using a consistency objective to align the clean actions implied by each mode. A velocity-weighted endpoint loss further strengthens endpoint supervision. During sampling, UniVEP first predicts endpoints at high noise and then switches to velocity prediction for low-noise refinement. Each step evaluates only the selected mode, retaining the inference cost of a single-target policy. Joint training also strengthens the endpoint predictor, allowing the policy to generate effective action~chunks~with~a~single~network~evaluation.

The resulting gains extend across simulated and real-robot manipulation. On RoboTwin Clean, UniVEP with StarVLA achieves \textbf{61.36\%} success in one evaluation, exceeding the strongest single-target baseline at sixteen evaluations (59.96\%). On LiLa-WAM, two evaluations achieve \textbf{85.96\%} success, surpassing the best sixteen-step single-target result (82.60\%) while cutting action-generation latency by \textbf{77.1\%}. On three real dual-arm tasks, mean success increases from 49.33\% to \textbf{58.67\%}.

Our contributions are threefold:
\begin{enumerate}[leftmargin=*,nosep,topsep=2pt]
    \item \textbf{Comprehensive comparison of prediction targets.} We systematically study V, X, and X--V across multiple VLA and WAM architectures, manipulation benchmarks, and inference budgets, revealing distinct performance trends as the sampling-step budget changes.
    \item \textbf{Understanding their complementary strengths.} Through action-space statistics and noise-conditioned prediction probes, we find that endpoint prediction is more stable at high noise, whereas velocity prediction is more effective for near-clean refinement.
    \item \textbf{UniVEP for efficient robot manipulation.} We unify endpoint and velocity prediction through joint supervision and noise-dependent switching. Across the policies we evaluate, UniVEP achieves strong manipulation performance with one or two network evaluations. At the same number of sampling steps, it matches \mbox{the inference FLOPs of a single-target policy.}
\end{enumerate}

\section{Related Work}
\label{sec:related_work}

\subsection{Prediction targets in generative models}
\label{sec:rw_visual_generation}

Diffusion and flow models predict noise, velocity, or clean data. Although algebraically related, these targets impose different learning objectives on finite networks~\citep{ho2020ddpm,lipman2023flowmatching,karras2022edm,ma2024sit,gagneux2026training,kumar2025loss,kim2026analyzinglearningdynamicslossprediction}. Recent studies connect target selection to data structure: JiT favors clean-image prediction~\citep{li2025jit}, AsymFlow uses full-rank data and low-rank noise~\citep{chen2026asymflow}, and MFlow blends targets along principal directions~\citep{mukhopadhyay2026mflow}. The influence of data dimensionality has also been investigated~\citep{jin2026dimension}.

\textbf{SC-Flow} is the closest methodological precedent~\citep{han2026scflow}. It jointly predicts velocity and endpoints within a shared network, enforces consistency between them, and switches from endpoint to velocity updates during sampling. Building on this formulation, UniVEP investigates observation-conditioned robot actions, where prediction quality must translate into closed-loop task success. We systematically examine prediction objectives across inference budgets and noise levels, together with velocity-weighted supervision and its interaction with pretrained action heads.

\subsection{Prediction objectives for robot actions}
\label{sec:rw_action_manifolds}

Diffusion policies generate action chunks via iterative denoising~\citep{chi2023diffusionpolicy,reuss2023beso,ze20243d,team2024octo,liu2025rdt}, while flow-based policies learn velocity fields~\citep{black2024pi0,braun2024rfmp,bjorck2025gr00t,shukor2025smolvla,pi2025pi05}. World-action models further couple actions with future visual features or video~\citep{yuan2026fastwam,yang2026lila,zhu2025uwm,shen2026videovla,ye2026world,song2026uniwam,liu2026cswam}. For dexterous video prediction, DexAC-WM~\citep{yuan2026not} preserves dimension-level action structure through per-dimension tokenization. Alternatively, ABot-M0 and multi-view Action Manifold Learning predict clean actions under velocity-space supervision~\citep{yang2026abot,xiao2026multiviewaml}, matching our X--V baseline except its bounded weighting. MM-ABC also reports gains from clean-action prediction~\citep{liang2026mmabc}.

We distinguish prediction targets from their training losses by comparing V, X, and X--V under matched training setups on StarVLA, FastWAM, and LiLa-WAM~\citep{starvla2026,yuan2026fastwam,yang2026lila}. Evaluations on LIBERO, LIBERO-Plus, and RoboTwin~\citep{liu2023libero,fei25libero-plus,chen2025robotwin} cover different inference budgets, perturbations, and scene randomization. We further investigate how velocity-oriented pretraining influences target selection in world-action models tied to a video backbone.

\subsection{Few-step generative control}
\label{sec:rw_fast_action}

Consistency Policy, ManiCM, and FlowPolicy accelerate robot action generation through consistency learning or distillation~\citep{prasad2024consistencypolicy,gao2024manicm,zhang2025flowpolicy}. One-Step Diffusion Policy and One-Step Flow Policy pursue single-evaluation control~\citep{wang2024onedp,li2026onestepflow}. Much Ado About Noising analyzes intermediate supervision, stochasticity, and iterative computation in generative robot control~\citep{pan2025noising}. Rather than distilling a separate student, we study prediction objectives within existing policies, distinguishing gains from joint training at one or two evaluations from those due to endpoint-to-velocity switching at larger budgets on both StarVLA and LiLa-WAM.

\section{Prediction Targets}
\label{sec:prediction_background}

\subsection{A Common Noise-to-Action Path}
\label{sec:targets}

Let $x\in\R^{HD}$ denote a clean action chunk with
$H$ action steps and $D$ dimensions per step.
The policy is conditioned on $c$, which contains the
available observations, language instruction, and
robot state. We sample
$\epsilon\sim\mathcal N(0,I_{HD})$ and interpolate
between noise and action along a linear path:
\begin{equation}
 z_t=(1-t)\epsilon+tx,\qquad v=x-\epsilon.
 \label{eq:path}
\end{equation}
Here $t\in[0,1]$ is the \emph{generation time},
not the time index of a robot action.
Sampling starts at $z_0=\epsilon$ and ends at $z_1=x$.
We write $\sigma=1-t$ for the remaining noise level.

A policy can predict either the path velocity $v$
or its clean endpoint $x$. The two outputs are
interchangeable \mbox{algebraically, at positive noise levels:}
\begin{equation}
 \begin{aligned}
 \hat x_{\mathrm V}&=z_t+\sigma\hat v,\\
 \hat v_{\mathrm X}&=\frac{\hat x-z_t}{\sigma},
 \qquad \sigma>0.
 \end{aligned}
 \label{eq:endpoint_from_v}
\end{equation}
Thus both predictions can drive the same flow sampler;
what changes is the quantity the network must learn.
\Cref{fig:prediction_formulations} illustrates the two outputs along the same path.

\subsection{Prediction Space vs. Loss Space}
\label{sec:target_objectives}

The network output and its training loss are separate
choices~\citep{li2025jit}. We compare the three recipes
in \cref{tab:target_formulations}. All recipes share the noisy input $(z_t,t,c)$ and sampling path.
The choice of prediction target and training loss defines the~three~cases~examined~in~this~work.

\begin{table}[!htbp]
 \centering
\begin{minipage}{.84\linewidth}
\centering
 \caption{\textbf{Three action prediction recipes.}
 The table omits the common expectation and
 $1/(HD)$ normalization.
 $\sigma=1-t$ and $\delta=0.05$ in all three recipes.}
 \label{tab:target_formulations}
 \definecolor{outV}{HTML}{D7EBD7}
 \definecolor{outX}{HTML}{FFF0D2}
 \papertablestyle\setlength{\tabcolsep}{8pt}
 \setlength{\fboxsep}{2pt}
 \renewcommand{\arraystretch}{1.05}
 \newcommand{\nout}[2]{\colorbox{#1}{\ensuremath{#2}}}
 \begin{fitwidth}
\begin{tabular}{@{}ccc@{}}
 \toprule
 \textbf{Recipe} & \textbf{Network output} & \textbf{Training loss} \\
 \midrule
 V & \nout{outV}{\hat v} & $\lVert\hat v-v\rVert_2^2$ \\
 X & \nout{outX}{\hat x} & $\lVert\hat x-x\rVert_2^2$ \\
 X--V & \nout{outX}{\hat x} &
 $\dfrac{\lVert\hat x-x\rVert_2^2}{\max(\sigma,\delta)^2}$ \\
 \bottomrule
 \end{tabular}
\end{fitwidth}
\end{minipage}
\end{table}

\noindent\textbf{V-prediction} directly regresses
the velocity $v$ and integrates its predicted field
$\hat v$. \textbf{X-prediction} directly regresses
the endpoint $x$ and converts $\hat x$ to a velocity
using \cref{eq:endpoint_from_v}.
\textbf{X--V} also predicts the endpoint, but
supervises it with a velocity-weighted loss.
For $\sigma\geq\delta$, its loss is exactly the
squared error of the converted velocity
$\hat v_{\mathrm X}$; the floor $\delta$ prevents
the weight from diverging near $t=1$.

Importantly, \textbf{X and X--V have identical
inference updates}. Their only difference is the
loss weighting during training.
At each fixed $(z_t,t,c)$, the ideal squared-error
predictors imply the same conditional mean endpoint
(\cref{eq:bayes_equivalence}); in practice, finite
networks trained with different objectives need not
perform equally well. The full objectives and
derivation are given in \cref{sec:app_prediction}.
The structure of the two regression targets provides a starting point for understanding these differences.

\Needspace{9\baselineskip}
\subsection{Structure of Action Targets}
\label{sec:geometry}
\finding{hiveViolet}{Compared with velocity targets and Gaussian noise, clean action chunks concentrate their variation~in~a~smaller~number~of~directions.}
\label{fnd:geometry}
We examine 2,000 RoboTwin Clean action chunks~\cite{liang2026hive}: 40 per task across 50 tasks, each with 448 coordinates ($H=32$, $D=14$).
Each action coordinate is standardized using its mean and standard deviation across all chunks. We draw matched Gaussian noise and form velocity targets $v=x-\epsilon$ in these coordinates.
Metrics are computed within each task and then averaged equally over tasks.

We use three measures to describe the target distributions.
For the centered sample matrix of each task, let $s_j$ be its singular values and $p_j=s_j^2/\sum_k s_k^2$ its spectral energy fractions.
\textbf{Participation ratio}, $\mathrm{PR}=1/\sum_jp_j^2$, measures the effective number of directions contributing to variation.
\textbf{99\% energy rank} counts the leading directions needed to retain 99\% of the squared singular values, including the weaker directions in the spectral tail.
\textbf{TwoNN} estimates local dimension from the ratio of second- to first-nearest-neighbor distances~\citep{facco2017twonn}; unlike the spectral measures, it uses local point spacing.

\begin{figure}[!htbp]
 \centering\includegraphics[width=0.64\linewidth]{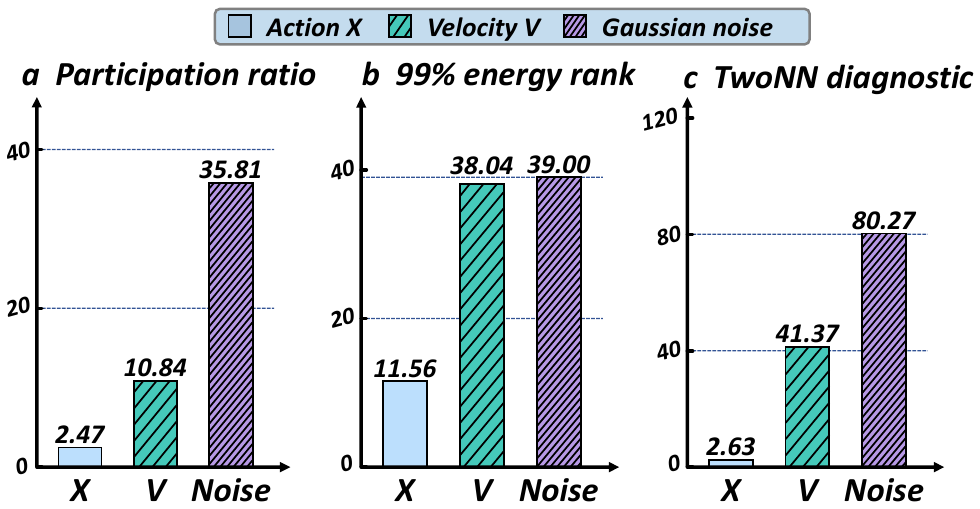}
 \caption{\textbf{Three views of the same structure.}
 Participation ratio, 99\% energy rank, and TwoNN for clean actions, velocity targets, and Gaussian noise.
 Bars are means over 50 tasks.}
 \label{fig:geometry}
 \label{fig:geometry_all_metrics}
\end{figure}

Clean actions have a mean PR of \textbf{2.47}, compared with 10.84 for velocity targets and 35.81 for noise.
Their mean 99\% energy ranks are \textbf{11.56}, 38.04, and 39.00, respectively; the corresponding TwoNN estimates are \textbf{2.63}, 41.37, and 80.27 (\cref{fig:geometry}).
All three measures show more concentrated structure in clean actions, providing a representation that endpoint prediction can exploit~\citep{li2025jit,yang2026abot}.
With 40 centered samples, the spectral rank ceiling is 39; TwoNN uses neighbor distances and has no such sample-matrix rank ceiling.
Full definitions and additional energy thresholds are in \cref{sec:app_geometry}.
These observations motivate testing how endpoint and velocity prediction perform under different supervision choices and inference budgets.

\FloatBarrier

\section{Analyzing Prediction Targets}
\label{sec:findings}

The target structure in \cref{sec:geometry} motivates a comparison of V, X, and X--V in closed-loop control.
We first measure task success across inference budgets, then probe how the learned predictions behave at different noise levels.
These observations guide the design of UniVEP in \cref{sec:method}.

\subsection{Evaluation Setup}
\label{sec:experimental_setup}
\label{sec:training_setup}

\paragraph{Benchmarks.}
We evaluate on three manipulation benchmarks covering a diverse set of manipulation paradigms:
\begin{itemize}[leftmargin=*,nosep,topsep=2pt]
    \item \textbf{LIBERO}~\citep{liu2023libero} evaluates single-arm manipulation across Spatial, Object, Goal, and Long suites. We report their unweighted mean success rate.
    \item \textbf{LIBERO-Plus}~\citep{fei25libero-plus} tests LIBERO-trained checkpoints under camera, robot, language, lighting, background, sensor-noise, and object-layout perturbations, without further training.
    \item \textbf{RoboTwin 2.0}~\citep{chen2025robotwin} contains 50 bimanual manipulation tasks. We evaluate both the demonstration-like \emph{Clean} split and the \emph{Randomized} split with altered object layouts and backgrounds. Inference-step sweeps use RoboTwin Clean.
\end{itemize}

\paragraph{Policy architectures and training.}
We study three representative flow-based action heads:
\begin{itemize}[leftmargin=*,nosep,topsep=2pt]
    \item \textbf{StarVLA}~\citep{starvla2026}: a flow-matching action head attached to a vision--language backbone. The vision--language and action learning rates are $(10^{-5},10^{-5})$ on LIBERO and $(10^{-5},10^{-4})$ on RoboTwin.
    \item \textbf{FastWAM}~\citep{yuan2026fastwam}: a world action model with velocity-pretrained components. It trains at $10^{-4}$ for 10 epochs on LIBERO and 5 epochs on RoboTwin. The scratch-action run drops only the ActionDiT initialization.
    \item \textbf{LiLa-WAM}~\citep{yang2026lila}: a world action model jointly predicting actions and future visual features. On RoboTwin, its Full configuration freezes DINO~\citep{simeoni2025dinov3} and trains for 12 epochs at $2\times10^{-4}$, then 4 epochs at $4\times10^{-5}$.
\end{itemize}

\paragraph{Controlled comparison.}
For each policy, we train V, X, and X--V with matched architecture, data, and update schedule, changing only the prediction target or its loss. On RoboTwin, StarVLA and FastWAM use 50 Clean demonstrations per task; LiLa-WAM Full uses 50 Clean and 500 Randomized demonstrations per task, so Randomized scores measure transfer only for the first two. \Cref{tab:recipes} summarizes device counts, per-device batches, update counts, and native inference budgets. \Cref{tab:baselines} reports native-budget results. \Cref{tab:nfe} reports three-seed means and standard deviations for RoboTwin Clean \smash[b]{\mbox{at each budget from one to sixteen evaluations.}}\vphantom{evaluations.}\looseness=1

\begin{table}[!htbp]
\centering
\begin{minipage}{.84\linewidth}
\captionsetup{font=footnotesize,skip=5pt}
\caption{\textbf{Training setup on NVIDIA A800 GPUs.} Batch counts demonstration samples per GPU per optimizer update; Steps are optimizer updates.}
\label{tab:recipes}
\fontsize{9}{10.5}\selectfont\setlength{\tabcolsep}{3pt}\renewcommand{\arraystretch}{1.05}
\begin{tabular*}{\linewidth}{@{\extracolsep{\fill}}llrrrr@{}}
\toprule
\hiveblue{Policy} & \hiveblue{Benchmark} & \hiveblue{GPUs} & \hiveblue{Batch/GPU} & \hiveblue{Steps} & \hiveblue{Native NFE}\\
\midrule
StarVLA & LIBERO & 4 & 16 & 120{,}000 & 4\\
StarVLA & RoboTwin & 8 & 8 & 75{,}000 & 4\\
FastWAM & LIBERO & 8 & 16 & 21{,}700 & 10\\
FastWAM & RoboTwin & 8 & 16 & 21{,}295 & 10\\
LiLa-WAM & LIBERO & 8 & 128 & 20{,}000 & 10\\
LiLa-WAM & RoboTwin & 8 & 128 & 95{,}632 & 10\\
\bottomrule
\end{tabular*}
\end{minipage}
\end{table}
\FloatBarrier

\subsection{Performance Across Inference Budgets}
\label{sec:init_findings}

\finding{hiveAccent}{On StarVLA and LiLa-WAM, X--V achieves the highest single-target success at one and two evaluations, while V gains more from additional evaluations than direct X.}
\label{fnd:architecture}

At their native inference budgets, X--V is the strongest single-target objective on LIBERO, LIBERO-Plus, and RoboTwin Clean for both StarVLA and LiLa-WAM (\cref{tab:baselines}). However, the relative performance of the objectives changes substantially with the number of evaluations.

\paragraph{Few-step generation favors X--V.}
With a single evaluation on RoboTwin Clean, X--V reaches \textbf{55.56\%} success on StarVLA, compared with 40.52\% for X and 35.92\% for V. The same pattern holds on LiLa-WAM: X--V reaches \textbf{80.12\%}, compared with 73.36\% for X and 67.40\% for V (\cref{tab:nfe}). Velocity-weighted endpoint supervision is therefore particularly effective under tight inference budgets.

\paragraph{Additional evaluations mainly benefit V.}
On StarVLA, V improves from 35.92\% at one evaluation to 53.84\% at sixteen, whereas X stays near 42\%. On LiLa-WAM, V increases from 67.40\% at one evaluation to 82.84\% at eight, approximately matching X--V at 82.56\%. In contrast, direct endpoint prediction changes little with additional evaluations. The step-sweep curves are provided in \cref{fig:nfe}.

\finding{hiveIndigo}{FastWAM depends on video–action coordination; mismatching the action prediction target can disrupt this coordination and reduce performance.}
\label{fnd:initialization}
At ten evaluations on RoboTwin Clean, pretrained FastWAM reaches 73.24\% success with V, 65.88\% with X, and 34.20\% with X--V (\cref{tab:baselines}). This reverses the relative ordering of V and X--V seen on StarVLA and LiLa-WAM at their native budgets. Both the video backbone and action expert in FastWAM are pretrained for velocity prediction~\citep{yuan2026fastwam}. Switching the action target to X or X--V asks the pretrained action expert to learn a different output while the video pathway keeps its velocity objective.

To isolate the role of action initialization, we train the action expert from scratch while retaining the pretrained video backbone and the joint video--action objective. At the same ten evaluations, X--V improves from 34.20\% to 64.48\% (\textbf{+30.28~pp}), X from 65.88\% to 69.20\% (+3.32~pp), and V changes from 73.24\% to 71.56\% ($-1.68$~pp). The gap between V and X--V narrows from 39.04 to 7.08~pp (\cref{tab:initialization}).

This difference is consistent with the action expert having to adapt a velocity-pretrained mapping to endpoint prediction, particularly under velocity-weighted endpoint supervision. V still leads at ten evaluations with either initialization. The full inference-budget sweep and additional controls are reported in \cref{sec:app_fastwam}.

\subsection{Prediction behavior across noise stages}
\label{sec:noise_findings}

Architecture and pretraining help explain why target preferences differ across policies. We next examine how these preferences change during generation, measuring initial-noise sensitivity and action recovery at different noise levels.

\paragraph{Shared probe protocol.}
We evaluate StarVLA, FastWAM, and LiLa-WAM Full on Clean and Randomized splits, using 2,004 matched observation anchors across 12 tasks per split and eight matched noise draws per anchor.
Conditions and checkpoints are held fixed. Each probe uses one action-network evaluation.
The initial-noise probe starts from Gaussian noise; the recovery probe corrupts recorded clean actions along \cref{eq:path} at $\sigma\in\{0.01,0.05,0.1,0.2,0.4,0.6,0.8,0.95,0.99\}$.
Measurements are averaged over episodes within each task and then equally across tasks.
The 95\% confidence intervals use hierarchical task/episode resampling with checkpoints and noise draws fixed (\cref{sec:app_probes}).

\finding{hiveTeal}{At high noise, endpoint prediction produces more stable actions across initial-noise draws.}
\label{fnd:noise_stability}\label{fnd:x_high_noise}

\begin{figure}[!htbp]
\centering\includegraphics[width=0.95\linewidth]{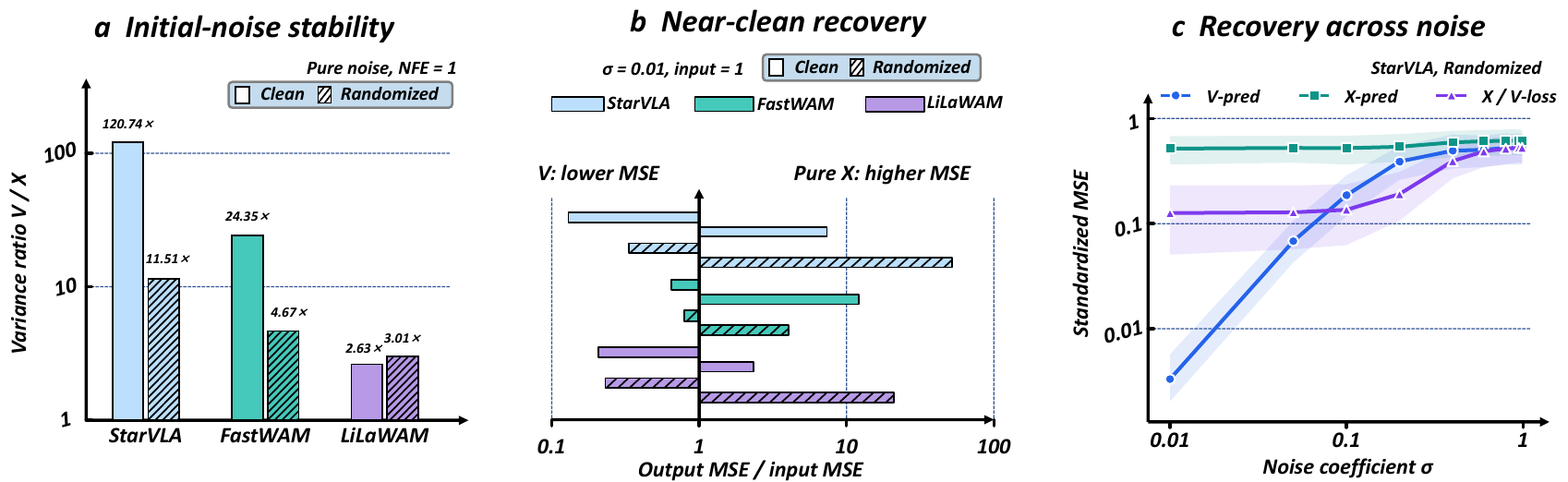}
\caption{\textbf{Prediction behavior across noise stages.}
\textbf{(a)} V/X output-variance ratios for generation from pure noise with one network call.
\textbf{(b)} Endpoint MSE divided by input MSE at $\sigma=0.01$, on a logarithmic axis. One marks the input baseline; V reduces error to its left, whereas Pure X amplifies error to its right.
\textbf{(c)} StarVLA Randomized recovery across noise levels. Error bars and shaded bands show 95\% task/episode bootstrap confidence intervals.}
\label{fig:noise}
\end{figure}

\paragraph{Initial-noise sensitivity.}
At $t=0$, X predicts a clean endpoint directly, whereas V reconstructs it as $\epsilon+\hat v$.
For each anchor, we measure the mean squared deviation of the eight standardized endpoint estimates from their mean, averaged over action coordinates.
This output variance measures how much the recovered action changes when only the initial noise changes.
\Cref{fig:noise}a reports V variance divided by X variance: values above one indicate greater noise sensitivity for V.

The ratios exceed one in all six settings: \textbf{120.74/11.51} on StarVLA Clean/Randomized, \textbf{24.35/4.67} on FastWAM, and \textbf{2.63/3.01} on LiLa-WAM Full.
Every 95\% interval lies above one (\cref{tab:one_step_stability}).
Thus, directly predicting the endpoint gives more consistent action estimates under the same observation and instruction.
This behavior complements the strong one-step endpoint results in \cref{tab:nfe}.

\label{sec:x_low_noise}\label{sec:low_noise}
\finding{hiveCyan}{Near clean actions, directly supervised endpoint prediction can increase error by replacing an already accurate action estimate.}
\label{fnd:x_low_noise}
We mix each recorded action with matched Gaussian noise at $\sigma=0.01$ and compare the network's endpoint estimate with that recorded action.
Error is measured as standardized endpoint MSE. Retaining the corrupted input provides the baseline.
\Cref{fig:noise}b normalizes each predictor's MSE by this input MSE, so ratios below one indicate recovery and ratios above one indicate error amplification.

Pure X lies above one in all six settings.
On StarVLA Randomized, its endpoint MSE is \textbf{52.1 times} the input error.
Four settings also have X denoising-gain intervals entirely below zero: StarVLA Randomized, both FastWAM splits, and LiLa-WAM Randomized (\cref{tab:noise_gain_intervals}).
Absolute MSE values are provided in \cref{tab:low_noise_evidence}.

At this stage, most of the action is already present in the input.
A direct endpoint predictor replaces the whole chunk, so preserving that information requires learned near-identity behavior.
\Cref{fig:noise}c shows that velocity-weighted endpoint supervision substantially lowers recovery error relative to Pure X on StarVLA Randomized.
This improvement connects the supervision choice to the low-noise recovery task, while retaining the endpoint output parameterization.

\label{sec:v_high_noise}
\finding{hiveAmber}{At high noise, velocity prediction must cancel input noise, and its prediction errors pass almost unchanged into the reconstructed endpoint.}
\label{fnd:v_high_noise}
From pure noise, V reconstructs an endpoint by adding its prediction to the random input.
Producing a consistent action therefore requires the velocity prediction to cancel the input's noise dependence.
Finding~\ref{fnd:geometry} motivates a linear-subspace interpretation: if clean actions occupy $d$ of the $HD$ action-space directions, X has zero targets in the remaining directions, whereas V must represent negative Gaussian noise there~\citep{feng2026diffusibility,chen2026asymflow}.
V thus represents both action structure and noise cancellation outside the action subspace.

The effect of an imperfect fit depends on the noise stage. Velocity error is multiplied by the remaining noise  $\sigma$ when converted to endpoint error. At high noise, $\sigma\approx1$, so cancellation errors pass almost unchanged into the recovered action; at low noise, the same reconstruction rule attenuates them.

\label{sec:v_low_noise}
\finding{hiveRose}{Near clean actions, velocity prediction reduces one-call endpoint recovery error. Its reconstruction adds \mbox{a correction scaled by the remaining noise.}}
\label{fnd:v_low_noise}\label{fnd:low_noise}
V retains the noisy action as its reconstruction base and adds the residual $\sigma\hat v$.
On the coefficient-consistent linear path~\citep{lee2026predictiontargets},
\begin{equation}
\hat x_{\mathrm V}-x=\sigma(\hat v-v).
\label{eq:low_noise_identity}
\end{equation}
Consequently, velocity MSE contributes to endpoint MSE with a factor of $\sigma^2$ in the same linear metric; at $\sigma=0.01$, this factor is $10^{-4}$.

The recovery probe measures whether the learned residual improves the input.
V reduces error in all six settings in \cref{fig:noise}b, where Pure X amplifies it.
On StarVLA Randomized, V leaves \textbf{0.334 times} the input MSE, a 66.6\% reduction.
All six V denoising-gain intervals are above zero, and all six paired V-minus-X MSE intervals are below zero (\cref{tab:noise_gain_intervals}).
These results show effective learned correction alongside the error attenuation in \cref{eq:low_noise_identity}. Across noise levels, \cref{fig:noise}c further shows lower recovery error for V near clean actions, even after velocity weighting improves endpoint prediction.

Together, the four findings identify complementary roles: endpoint prediction provides stable recovery from high noise, while velocity prediction refines actions at low noise. This division of roles motivates the noise-dependent sampler used by UniVEP.

\begin{figure}[!htbp]
    \centering
    \includegraphics[width=\linewidth]{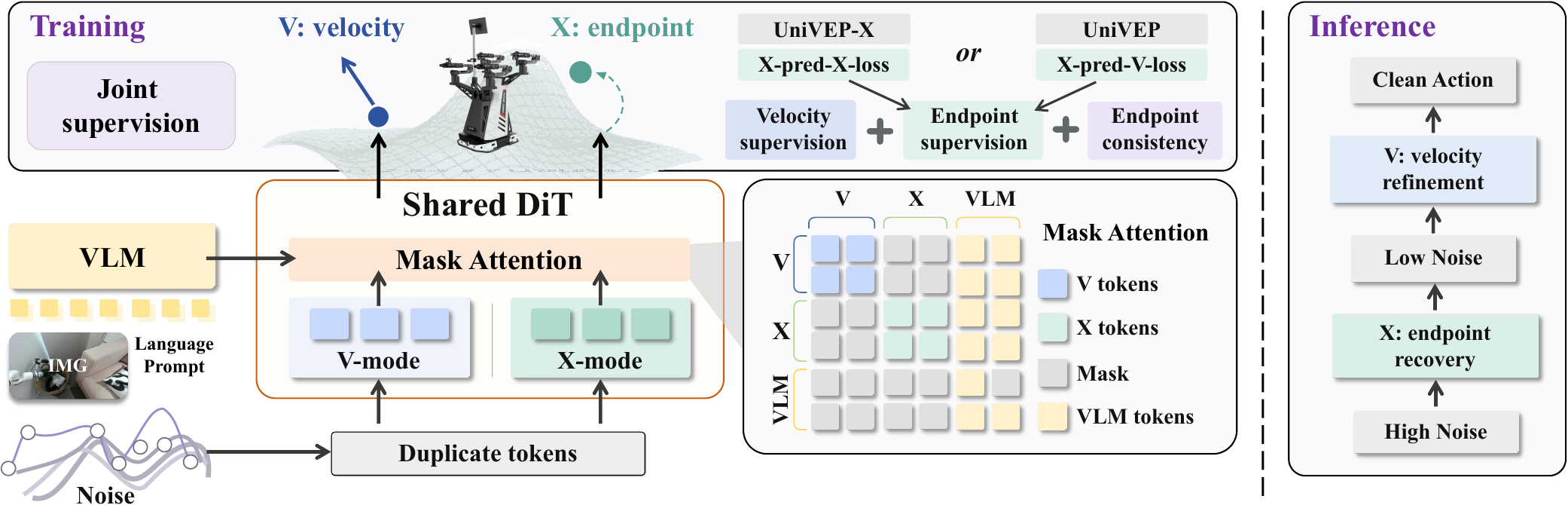}
    \caption{\textbf{UniVEP: joint training and noise-dependent inference.}
    \textbf{Training (left):} The same noisy action chunk and VLM conditions train velocity (V) and endpoint (X) in a shared DiT. The mask separates the two action streams while both keep shared conditioning tokens. Both are supervised, and consistency aligns their implied clean actions. UniVEP uses a bounded velocity-weighted endpoint loss; UniVEP--X uses an unweighted one.
    \textbf{Inference (right):} Each step evaluates only the selected mode: X at high noise ($\sigma\geq\sigma^*$), and V at low noise ($\sigma<\sigma^*$).}
    \label{fig:overview}
\end{figure}

\section{UniVEP}
\label{sec:method}

The findings in \cref{sec:noise_findings} suggest different roles for endpoint and velocity prediction during action generation. UniVEP brings both modes into a single action policy: \textbf{training supervises both predictions, while inference evaluates only the mode selected for the current noise level} (\cref{fig:overview}).

\subsection{A Shared Predictor with Two Modes}
\label{sec:method_parameterization}

Let $c$ denote the observation, instruction, and optional robot-state features supplied by the original policy. Given the noisy action chunk $z_t$ and time $t$ from \cref{eq:path}, a mode embedding selects one of two outputs from the network~\citep{han2026scflow}:
\begin{equation}
    \hat v_\theta=f_\theta(z_t,t,c,0),\qquad
    \hat x_\theta=f_\theta(z_t,t,c,1).
    \label{eq:mode_network}
\end{equation}
Here $0$ denotes V and $1$ denotes X. Both predictions have shape $H\times D$ and share the action transformer; only the mode embedding differs.

Both modes see the same $(z_t,t,c)$. StarVLA and LiLa-WAM duplicate tokens after the image encoder; FastWAM reuses its video tokens with two masked streams (\cref{sec:app_univep}).

\subsection{Joint Supervision and Consistency}
\label{sec:method_training}

UniVEP combines velocity regression, endpoint regression, and a consistency term that aligns the \mbox{clean action estimates from both modes.}
\begingroup
\abovedisplayskip=\abovedisplayshortskip
\belowdisplayskip=\belowdisplayshortskip
\begin{equation}
\begin{aligned}
\mathcal L_{\mathrm{UniVEP}}
 &=\E_{(c,x),\epsilon,t}\!\left[\ell_{\mathrm V}+\ell_{\mathrm X}+\ell_{\mathrm C}\right],\\
\ell_{\mathrm V}&=\frac{1}{HD}\norm{\hat v_\theta-v}_2^2,\\
\ell_{\mathrm X}&=\frac{w_X(t)}{HD}\norm{\hat x_\theta-x}_2^2,\\
\ell_{\mathrm C}&=\frac{1}{HD}\norm{\hat x_\theta-(z_t+\sigma\hat v_\theta)}_2^2.
\end{aligned}
\label{eq:ufo}
\end{equation}
\endgroup
The first two terms supervise their respective targets. The consistency term couples the modes by matching the endpoint predicted directly by X to the endpoint reconstructed from V. All three losses use the same noise--time sample, and gradients from $\ell_{\mathrm C}$ update the shared parameters through both modes.
The two joint-training variants differ only in endpoint weighting:
\begin{equation}
 w_X(t)=
 \begin{cases}
    1, & \text{UniVEP--X},\\[2pt]
    \dfrac{1}{\max(1-t,\delta)^2}, & \text{UniVEP},\quad\delta=0.05.
 \end{cases}
 \label{eq:architecture_supervision_rule}
\end{equation}
For $1-t\geq\delta$, the weighted endpoint term is equivalent to the velocity-space error of the endpoint prediction (\cref{sec:target_objectives}); the floor prevents excessively large weights near $t=1$. \textbf{Both variants retain V supervision and the same consistency loss.} StarVLA and LiLa-WAM use the weighted variant; we report both for FastWAM, whose velocity-pretrained components respond differently to endpoint weighting (\cref{sec:app_fastwam}). The original video or future-feature objectives, time sampling, and optimizer schedules are otherwise preserved (\cref{sec:training_setup}).

\subsection{Noise-Dependent Sampling}
\label{sec:method_sampling}

At a noisy state $z_t$, the modes provide two estimates of the clean action: $\hat x_{\mathrm X}=\hat x_\theta$ and $\hat x_{\mathrm V}=z_t+\sigma\hat v_\theta$. To compare them in the same space, define their endpoint risk as
\begin{equation}
 \mathcal R_m(\sigma)
 =\frac{1}{HD}\E\!\left[\norm{\hat x_m-x}_2^2\mid\sigma\right],
 \quad m\in\{\mathrm X,\mathrm V\}.
 \label{eq:endpoint_risk}
\end{equation}
The velocity-induced endpoint error is scaled by $\sigma$ (\cref{eq:low_noise_identity}), making V particularly suitable for near-clean correction. Combined with the high-noise stability of X observed in \cref{sec:noise_findings}, this motivates an X-first, V-later sampling schedule for the grids below.

For a grid $0=t_0<\cdots<t_N=1$, let $h_i=t_{i+1}-t_i$ and $\sigma_i=1-t_i$. UniVEP takes the Euler update using just one mode at each step:
\begin{equation}
 z_{i+1}=
 \begin{cases}
 z_i+\dfrac{h_i}{\sigma_i}(\hat x_\theta-z_i),
       & \sigma_i\geq\sigma^*,\\[6pt]
 z_i+h_i\hat v_\theta,
       & \sigma_i<\sigma^*.
 \end{cases}
 \label{eq:mixed_sampling}
\end{equation}
The predictions in each branch are evaluated at $(z_i,t_i,c)$. We use a fixed $\sigma^*=0.5$ and ablate this choice in \cref{tab:only_v}. The consistency term encourages the two possible updates to agree at the same state; their difference is proportional to the residual in $\ell_{\mathrm C}$ (\cref{eq:mode_update_agreement}). \Cref{eq:local_update_risk} shows how endpoint error~propagates~into~a~single~sampling~step.

\paragraph{One or two evaluations.}
With $N=1$, the endpoint branch directly outputs $z_1=\hat x_\theta$. With $N=2$, both steps also use X on the grids evaluated here. Consequently, improvements at these budgets come from \emph{jointly trained endpoint prediction}, not from inference-time switching. At larger budgets, later evaluations can use V for refinement. Each step still invokes the shared DiT only once, matching the single-target inference FLOPs at the same number of evaluations. Step allocations and pseudocode are in \cref{sec:app_univep,alg:sample}; each action chunk is executed with the host policy's original controller, and the executed horizon stays fixed.

\section{Experiments}
\label{sec:experiments}

We evaluate UniVEP along four dimensions: task success across
manipulation benchmarks, performance under limited sampling budgets,
the contribution of individual training and sampling choices, and
wall-clock inference cost. Evaluation protocols and training budgets
follow the host policies (\cref{sec:experimental_setup}).

\subsection{Performance Across Benchmarks}
\label{sec:joint_nfe}
\label{sec:case_joint_nfe}

\finding{hiveAccent}{UniVEP improves LIBERO-Plus and achieves the best RoboTwin results for StarVLA and LiLa-WAM among the compared recipes.}
\label{fnd:method_robustness}

\begin{table}[!htbp]
\centering
\begin{minipage}{.78\linewidth}
\centering
\caption{\textbf{Native-budget manipulation success (\%).}
StarVLA uses 4 NFE; FastWAM and LiLa-WAM use 10.
Plus is LIBERO-Plus; Clean/Rand. are RoboTwin splits.
For RoboTwin, only LiLa-WAM Full uses Randomized training data.
Bold marks the best score within each policy and benchmark.
StarVLA and LiLa-WAM Clean entries are three-seed means (\cref{tab:nfe}).}
\label{tab:baselines}
\label{tab:ufo_libero}
\papertablestyle\fontsize{8}{9.3}\selectfont\setlength{\tabcolsep}{3pt}
\begin{fitwidth}
\begin{tabular*}{\linewidth}{@{\extracolsep{\fill}}lrrrr@{}}
\toprule
\thdr
\textbf{Recipe} & \textbf{LIBERO} & \textbf{Plus} & \textbf{Clean} & \textbf{Rand.}\\
\midrule
\tgroup
\multicolumn{5}{@{}l}{\textbf{StarVLA}}\\
V & 97.45 & 75.70 & \arxivthird{43.04} & 7.04\\
X & \arxivsecond{98.15} & \arxivthird{78.20} & 42.80 & \arxivthird{8.16} \\
X--V & \arxivfirst{\textbf{98.45}} & \arxivsecond{79.60} & \arxivsecond{58.02} & \arxivsecond{10.84} \\
\tmethod
\textbf{\method{}} & \arxivthird{97.75} & \arxivfirst{\textbf{80.40}} & \arxivfirst{\textbf{61.56}} & \arxivfirst{\textbf{13.68}} \\
\midrule
\tgroup
\multicolumn{5}{@{}l}{\textbf{FastWAM}}\\
V & 96.20 & 50.40 & \arxivfirst{\textbf{73.24}} & 2.72\\
X & 95.45 & 50.30 & \arxivthird{65.88} & 2.00\\
X--V & \arxivthird{96.80} & 47.30 & 34.20 & 1.00\\
\tmethod
\textbf{\method{}} & \arxivsecond{97.25} & \arxivthird{52.40} & 57.00 & \arxivsecond{3.22} \\
\method{} (scratch) & \arxivfirst{\textbf{98.38}} & \arxivfirst{\textbf{66.00}} & 56.12 & \arxivthird{3.16} \\
\method{}--X & 96.25 & \arxivsecond{54.70} & \arxivsecond{66.36} & \arxivfirst{\textbf{3.46}} \\
\midrule
\tgroup
\multicolumn{5}{@{}l}{\textbf{LiLa-WAM Full}}\\
V & 96.10 & 52.30 & 83.24 & 76.32\\
X & 96.00 & 54.00 & 75.04 & 73.96\\
X--V & \arxivthird{96.65} & \arxivsecond{55.40} & \arxivthird{83.48} & \arxivthird{78.28} \\
\method{}--X & \arxivsecond{96.70} & \arxivthird{55.25} & \arxivsecond{83.72} & \arxivsecond{81.24} \\
\tmethod
\textbf{\method{}} & \arxivfirst{\textbf{97.10}} & \arxivfirst{\textbf{58.40}} & \arxivfirst{\textbf{87.04}} & \arxivfirst{\textbf{83.04}} \\
\bottomrule
\end{tabular*}
\end{fitwidth}
\end{minipage}
\end{table}

\Cref{tab:baselines} reports results at each policy's standard
sampling budget. \textbf{On RoboTwin Clean}, UniVEP improves StarVLA
from 58.02\% (X--V) to 61.56\% and LiLa-WAM from 83.48\% to
87.04\%. The gains extend to more varied scenes: StarVLA reaches
80.40\% on LIBERO-Plus and 13.68\% on RoboTwin Randomized,
compared with 79.60\% and 10.84\% for X--V. With LiLa-WAM Full,
UniVEP improves LIBERO-Plus from 55.40\% to 58.40\%, and RoboTwin
Randomized from 78.28\% to 83.04\%. LiLa-WAM Full trains on Randomized demonstrations. StarVLA evaluates the Randomized split using a policy~trained~on~Clean~demonstrations.\looseness=1

\textbf{Standard LIBERO is near saturation.} All three policies already
achieve high success rates, and UniVEP is not the top StarVLA recipe
(97.75\% versus 98.45\% for X--V). The larger differences emerge on
the perturbed or bimanual evaluations rather than on this saturated
benchmark.

\textbf{FastWAM is an important exception.} On RoboTwin Clean,
its velocity-pretrained configuration favors V (73.24\%) over
UniVEP (57.00\%) and UniVEP--X (66.36\%). Removing only the pretrained
action-head initialization raises the X--V baseline from 34.20\% to
64.48\% at 10 NFE (\cref{sec:app_fastwam}), indicating substantial
sensitivity to initialization. Conversely, scratch-action UniVEP
improves LIBERO-Plus to 66.00\%. We therefore report FastWAM
separately rather than treating improvements as universal.

\subsection{Few-Step Action Generation}
\label{sec:few_step_results}

\finding{hiveTeal}{UniVEP inherits X-prediction's few-step efficiency, generating high-quality actions for strong and stable control with one or two evaluations.}
\label{fnd:joint_few_step}
\label{fnd:few_step}

\begin{table}[!htbp]
\centering
\begin{minipage}{.84\linewidth}
\centering
\caption{\textbf{RoboTwin Clean success (\%) versus NFE.}
Mean $\pm$ standard deviation across three training seeds over 50 tasks.
Each recipe uses the same trained weights across NFE.
Bold marks the best score within each policy and NFE.}
\label{tab:nfe}
\papertablestyle\fontsize{9}{10.5}\selectfont\setlength{\tabcolsep}{3pt}
\begin{fitwidth}
\begin{tabular*}{\linewidth}{@{\extracolsep{\fill}}lccccc@{}}
\toprule
\thdr
\textbf{NFE} & \textbf{1} & \textbf{2} & \textbf{4} & \textbf{8} & \textbf{16}\\
\midrule
\tgroup
\multicolumn{6}{@{}l}{\textbf{StarVLA}}\\
V & $35.92\pm0.82$ & \arxivthird{$49.32\pm2.94$} & \arxivthird{$43.04\pm0.24$} & \arxivthird{$51.32\pm1.12$} & \arxivthird{$53.84\pm2.54$} \\
X & \arxivthird{$40.52\pm0.24$} & $41.52\pm1.04$ & $42.80\pm1.36$ & $42.16\pm0.88$ & $42.52\pm1.18$\\
X--V & \arxivsecond{$55.56\pm1.28$} & \arxivsecond{$59.60\pm2.82$} & \arxivsecond{$58.02\pm2.36$} & \arxivsecond{$58.32\pm3.34$} & \arxivsecond{$59.96\pm2.80$} \\
\tmethod
\textbf{\method{}} & \arxivfirst{$\mathbf{61.36\pm0.26}$} & \arxivfirst{$\mathbf{63.24\pm0.98}$} & \arxivfirst{$\mathbf{61.56\pm2.42}$} & \arxivfirst{$\mathbf{64.40\pm2.36}$} & \arxivfirst{$\mathbf{66.88\pm2.32}$} \\
\midrule
\tgroup
\multicolumn{6}{@{}l}{\textbf{LiLa-WAM Full}}\\
V & $67.40\pm3.58$ & $80.28\pm3.62$ & $81.52\pm2.20$ & \arxivthird{$82.84\pm1.42$} & $81.92\pm0.18$\\
X & $73.36\pm2.88$ & $72.52\pm2.36$ & $73.88\pm1.72$ & $74.48\pm0.32$ & $75.32\pm2.64$\\
X--V & \arxivsecond{$80.12\pm2.22$} & \arxivsecond{$82.68\pm0.60$} & \arxivthird{$82.08\pm3.36$} & $82.56\pm0.44$ & \arxivthird{$82.60\pm2.86$} \\
\method{}--X & \arxivthird{$75.64\pm2.36$} & \arxivthird{$82.36\pm1.64$} & \arxivsecond{$85.60\pm0.58$} & \arxivsecond{$84.64\pm0.66$} & \arxivsecond{$84.48\pm1.02$} \\
\tmethod
\textbf{\method{}} & \arxivfirst{$\mathbf{81.88\pm2.80}$} & \arxivfirst{$\mathbf{85.96\pm2.76}$} & \arxivfirst{$\mathbf{86.08\pm0.20}$} & \arxivfirst{$\mathbf{86.52\pm3.26}$} & \arxivfirst{$\mathbf{86.72\pm0.32}$} \\
\bottomrule
\end{tabular*}
\end{fitwidth}
\end{minipage}
\end{table}

The strongest gains occur when the sampling budget is smallest
(\cref{tab:nfe}). \textbf{With one StarVLA evaluation}, UniVEP reaches
61.36\%, exceeding not only one-step X--V (55.56\%) but also its
16-step result (59.96\%). With \textbf{two LiLa-WAM evaluations},
UniVEP reaches 85.96\%, exceeding the best single-target result at
8, 10, or 16 evaluations (82.84\%, 83.48\%, and 82.60\%, respectively).
Even one-step LiLa-WAM UniVEP reaches 81.88\%, above one-step X--V
at 80.12\%. The gains are not limited to the Clean split: LiLa-WAM
UniVEP reaches 82.60\% with two evaluations on Randomized scenes,
versus 78.28\% for ten-step X--V (\cref{tab:app_lila_randomized}) across all fifty tasks there.

Importantly, the \textbf{one- and two-step samplers evaluate only X}
on our sampling grids (\cref{sec:method_sampling}). Their gains
therefore arise from the jointly trained endpoint predictor, not
from switching modes during inference. At larger budgets, switching
also enables refinement: StarVLA UniVEP reaches 66.88\% at 16 NFE,
while LiLa-WAM maintains 86--87\% success from 4 to 16 NFE.
The corresponding measured latency savings are quantified in
\cref{sec:compute_controls}.

\subsection{Training Ablations}
\label{sec:training_ablation}

\finding{hiveRose}{Joint X/V supervision and endpoint consistency both improve control.}
\label{fnd:joint_endpoint}

\begin{table}[!htbp]
\centering
\begin{minipage}{.82\linewidth}
\captionsetup{font=footnotesize,skip=5pt}
\caption{\textbf{Training-objective ablations on RoboTwin Clean.} Full uses UniVEP for StarVLA and LiLa-WAM, and UniVEP--X for FastWAM. In (a), the single-target baseline is X--V for StarVLA and LiLa-WAM, and X for FastWAM. Success rates are in percent, and Gain is the difference in percentage points.}
\label{tab:endpoint_training_ablation}
\fontsize{9}{10.5}\selectfont\setlength{\tabcolsep}{3pt}\renewcommand{\arraystretch}{1.05}
\begin{fitwidth}
\begin{tabular*}{\linewidth}{@{\extracolsep{\fill}}llrrrr@{}}
\toprule
\hiveblue{Removed component} & \hiveblue{Policy} & \hiveblue{NFE} & \hiveblue{Full} & \hiveblue{w/o component} & \hiveblue{Gain (pp)}\\
\midrule
(a) Joint X/V training & StarVLA & 1 & \rankfirst{61.36} & 55.56 & \hivebad{$+5.80$}\\
 & StarVLA & 2 & \rankfirst{63.24} & 59.60 & \hivebad{$+3.64$}\\
 & LiLa-WAM & 1 & \rankfirst{81.88} & 80.12 & \hivebad{$+1.76$}\\
 & LiLa-WAM & 2 & \rankfirst{85.96} & 82.68 & \hivebad{$+3.28$}\\
 & FastWAM & 1 & \rankfirst{66.08} & 59.64 & \hivebad{$+6.44$}\\
 & FastWAM & 2 & \rankfirst{66.16} & 61.76 & \hivebad{$+4.40$}\\
 & FastWAM & 4 & \rankfirst{66.04} & 62.80 & \hivebad{$+3.24$}\\
\midrule
(b) Endpoint consistency & StarVLA & 4 & \rankfirst{61.56} & 57.52 & \hivebad{$+4.04$}\\
 & LiLa-WAM & 10 & \rankfirst{87.04} & 83.68 & \hivebad{$+3.36$}\\
 & FastWAM & 10 & \rankfirst{66.36} & 59.28 & \hivebad{$+7.08$}\\
\bottomrule
\end{tabular*}
\end{fitwidth}
\end{minipage}
\end{table}

\paragraph{Joint training improves the endpoint itself.}
All three policies use only X at one and two evaluations, and FastWAM
also uses only X at four evaluations on its native grid.
Joint training improves success over the corresponding single-target
recipe in every setting in \cref{tab:endpoint_training_ablation}(a).
For FastWAM, the gains over X are 6.44, 4.40, and 3.24~pp at one, two,
and four evaluations, respectively.

\paragraph{Consistency matters beyond dual supervision.}
With both modes supervised and the X$\to$V sampler held fixed,
endpoint consistency raises native-budget success by 4.04~pp on
StarVLA, 3.36~pp on LiLa-WAM, and 7.08~pp on FastWAM
(\cref{tab:endpoint_training_ablation}(b)). Per-task results are
in \cref{tab:app_consistency_ablation}.

\finding{hiveAmber}{Velocity-weighted endpoint supervision benefits LiLa-WAM across inference budgets, while FastWAM on RoboTwin Clean benefits from greater emphasis on its pretrained velocity objective.}
\label{fnd:joint_design}
On LiLa-WAM, replacing the bounded velocity-weighted endpoint
loss with a flat loss (UniVEP--X) reduces RoboTwin Clean success from
87.04\% to 83.72\% at 10 NFE, and from 81.88\% to 75.64\% at one NFE
(\cref{tab:nfe}). The Randomized score at 10 NFE decreases from
83.04\% to 81.24\% (\cref{tab:app_lila_randomized}). FastWAM displays
the opposite preference on RoboTwin Clean (66.36\% for UniVEP--X
versus 57.00\% for weighted UniVEP). This pattern is consistent with the initialization results in \cref{sec:app_fastwam}, where the pretrained~video~pathway~is~held~fixed.

\subsection{Sampling Ablations}
\label{sec:ablations}
\label{sec:route_ablation}

\finding{hiveIndigo}{Reversing X to V order reduces native-NFE success across all three policies; both sampling phases improve LiLa-WAM and FastWAM control.}
\label{fnd:x_first_route}

\begin{table}[!htbp]
\centering
\begin{minipage}{.84\linewidth}
\captionsetup{font=footnotesize,skip=5pt}
\caption{\textbf{RoboTwin Clean inference ablations.} StarVLA and LiLa-WAM use UniVEP, and FastWAM uses UniVEP--X. Each comparison holds the trained weights fixed.}
\label{tab:only_v}\label{tab:switch_threshold}
\fontsize{9}{10.5}\selectfont\setlength{\tabcolsep}{3pt}\renewcommand{\arraystretch}{1.05}
\begin{tabular*}{\linewidth}{@{\extracolsep{\fill}}lrrrr@{}}
\toprule
\multicolumn{5}{@{}l}{\textcolor{hiveAccent}{\textbf{(a) Sampling order at native NFE}}}\\
\hiveblue{Policy} & \hiveblue{NFE} & \hiveblue{$X\!\to\!V$} & \hiveblue{$V\!\to\!X$} & \hiveblue{Gain (pp)}\\
\midrule
StarVLA & 4 & \rankfirst{61.56} & 58.88 & \hivebad{$+2.68$}\\
LiLa-WAM & 10 & \rankfirst{87.04} & 84.32 & \hivebad{$+2.72$}\\
FastWAM & 10 & \rankfirst{66.36} & 63.80 & \hivebad{$+2.56$}\\
\bottomrule
\end{tabular*}
\par\smallskip
\begin{tabular*}{\linewidth}{@{\extracolsep{\fill}}lrrrrrr@{}}
\toprule
\multicolumn{7}{@{}l}{\textcolor{hiveRose}{\textbf{(b) Contributions of the two sampling phases}}}\\
\hiveblue{Policy} & \hiveblue{NFE} & \hiveblue{$X\!\to\!V$} & \hiveblue{X-only} & \hiveblue{Gain (pp)} & \hiveblue{V-only} & \hiveblue{Gain (pp)}\\
\midrule
LiLa-WAM & 4 & \rankfirst{86.08} & 84.40 & \hivebad{$+1.68$} & 82.68 & \hivebad{$+3.40$}\\
 & 8 & \rankfirst{86.52} & 83.52 & \hivebad{$+3.00$} & 83.56 & \hivebad{$+2.96$}\\
 & 10 & \rankfirst{87.04} & 82.88 & \hivebad{$+4.16$} & 82.68 & \hivebad{$+4.36$}\\
 & 16 & \rankfirst{86.72} & 82.80 & \hivebad{$+3.92$} & 82.64 & \hivebad{$+4.08$}\\
\addlinespace[2pt]
FastWAM & 8 & \rankfirst{66.28} & 63.20 & \hivebad{$+3.08$} & 64.04 & \hivebad{$+2.24$}\\
 & 10 & \rankfirst{66.36} & 63.16 & \hivebad{$+3.20$} & 64.48 & \hivebad{$+1.88$}\\
 & 16 & \rankfirst{66.32} & 63.92 & \hivebad{$+2.40$} & 64.96 & \hivebad{$+1.36$}\\
\bottomrule
\end{tabular*}
\par\smallskip
\begin{tabular*}{\linewidth}{@{\extracolsep{\fill}}lrrrrr@{}}
\toprule
\multicolumn{6}{@{}l}{\textcolor{hiveAmber}{\textbf{(c) Switching threshold: StarVLA, NFE=16}}}\\
\hiveblue{$\sigma^*$} & \hiveblue{0.750} & \hiveblue{0.625} & \hiveblue{0.500} & \hiveblue{0.375} & \hiveblue{0.250}\\
\midrule
X$\to$V allocation & 5X$\to$11V & 7X$\to$9V & \textbf{9X$\to$7V} & 11X$\to$5V & 13X$\to$3V\\
Success (\%) & 55.62 & 63.24 & \rankfirst{66.88} & 61.78 & 53.24\\
\bottomrule
\end{tabular*}
\par\smallskip\footnotesize Gains in (a,b) are the success-rate differences between $X\!\to\!V$ and each alternative.
\end{minipage}
\end{table}

\paragraph{Both modes contribute at inference.}
At native inference budgets, X$\to$V outperforms V$\to$X by 2.68~pp
on StarVLA, 2.72~pp on LiLa-WAM, and 2.56~pp on FastWAM
(\cref{tab:only_v}(a)). With the same weights, X$\to$V also beats
both single-mode schedules at every tested budget for LiLa-WAM and
FastWAM (\cref{tab:only_v}(b)). At 10 NFE, FastWAM reaches 66.36\%
with X$\to$V, compared with 63.16\% for X-only and 64.48\% for V-only.
Per-task sampling results are reported in
\cref{tab:app_lila_reverse,tab:app_route_ablation}.

\paragraph{The switch location matters.}
On StarVLA at 16 NFE, the tested switching thresholds form a clear performance peak at
$\sigma^*=0.5$ (66.88\%), corresponding to nine X updates followed
by seven V updates. Switching earlier ($\sigma^*=0.75$) yields 55.62\%,
whereas switching later ($\sigma^*=0.25$) yields 53.24\%
(\cref{tab:only_v}(c)). We use the threshold selected by this task-success sweep.

\subsection{Inference Efficiency}
\label{sec:compute_controls}

\finding{hiveTeal}{UniVEP achieves strong control with unchanged counted FLOPs and comparable latency~at~matched~inference~budgets.}
\label{fnd:compute_efficiency}

\begin{table}[!htbp]
\centering
\begin{minipage}{.84\linewidth}
\caption{\textbf{RoboTwin inference latency (ms/chunk) by NFE.}
Timing runs on one RTX 4090 at batch size 1 through delivery of the action chunk to the CPU.
Each recipe uses fixed weights across NFE.}
\label{tab:compute_nfe_latency}
\papertablestyle\setlength{\tabcolsep}{3pt}
\begin{tabularx}{\linewidth}{@{}PCCCCCC@{}}
\toprule
\hiveblue{Policy / recipe} & \hiveblue{1} & \hiveblue{2} & \hiveblue{4} & \hiveblue{8} & \hiveblue{10} & \hiveblue{16}\\
\midrule
\tgroup
\multicolumn{7}{@{}l}{\textbf{StarVLA}}\\
X & 60.16 & 68.61 & 85.13 & 119.27 & 135.50 & 186.99\\
V & 59.29 & 68.60 & 85.35 & 120.17 & 137.31 & 187.74\\
X--V & 59.68 & 68.71 & 86.42 & 122.86 & 139.09 & 191.32\\
\method{} & 61.59 & 70.17 & 86.94 & 122.98 & 139.15 & 193.10\\
\midrule
\tgroup
\multicolumn{7}{@{}l}{\textbf{LiLa-WAM Full}}\\
X & 17.37 & 23.01 & 33.86 & 56.23 & 67.50 & 101.01\\
V & 17.72 & 24.00 & 34.23 & 56.90 & 68.25 & 101.09\\
X--V & 17.34 & 22.76 & 34.16 & 56.35 & 67.75 & 100.98\\
\method{}--X & 17.53 & 23.69 & 35.10 & 57.28 & 69.48 & 102.58\\
\method{} & 17.66 & 23.16 & 35.30 & 57.15 & 69.07 & 102.93\\
\midrule
\tgroup
\multicolumn{7}{@{}l}{\textbf{FastWAM}}\\
X & 101.72 & 134.32 & 205.61 & 303.91 & 362.80 & 543.40\\
V & 101.46 & 131.51 & 190.11 & 306.16 & 362.13 & 542.45\\
X--V & 100.62 & 128.58 & 198.94 & 317.18 & 361.24 & 545.03\\
\method{}--X & 103.06 & 132.46 & 191.81 & 310.99 & 368.17 & 548.01\\
\method{} & 99.97 & 128.77 & 186.87 & 302.44 & 360.89 & 535.58\\
\bottomrule
\end{tabularx}
\end{minipage}
\end{table}

At a matched policy and NFE, UniVEP evaluates one prediction mode per step
and has the same counted inference FLOPs as the single-target recipes.
At the native budgets, the cost per action chunk is 2.411~TFLOPs
for StarVLA (4 NFE), 0.466 for LiLa-WAM (10 NFE), and 4.041 for FastWAM (10 NFE).
At the native budgets in \cref{tab:compute_nfe_latency}, each UniVEP variant's measured latency stays within approximately 3\% of that of the corresponding single-target recipes.

Reducing the number of evaluations yields larger savings. One-step StarVLA UniVEP
(61.36\%, 61.59\,ms) exceeds 16-step X--V
(59.96\%, 191.32\,ms), a \textbf{67.8\% latency reduction}.
Two-step LiLa-WAM UniVEP (85.96\%, 23.16\,ms) exceeds the
10-step X--V baseline (83.48\%, 67.75\,ms) at
\textbf{65.8\% lower latency}; against the 16-step X--V baseline
(82.60\%, 100.98\,ms), the reduction is \textbf{77.1\%}.

Joint training has a separate training-time cost: for StarVLA on one
A800, it adds 6.93\% counted FLOPs and 15.57\% per-update time over V
(\cref{tab:training_cost}). Timing protocols and additional controls
are in \cref{sec:app_compute}.

\subsection{Real-World Bimanual Manipulation}
\label{sec:real_robot}

We evaluate UniVEP on a 14-DoF ARX A7 dual-arm platform
equipped with an egocentric Intel RealSense D405 depth camera.
The hardware configuration and workspace are shown in
\cref{fig:real_robot}(a). The policy takes a
$224\times224$ egocentric image, a language instruction,
and a 14-dimensional joint state as input, and predicts
16-step chunks of absolute joint positions. We consider three manipulation tasks that probe sequential
object interaction, bimanual coordination, and precise
gripper alignment:

\begin{figure}[!htbp]
\centering
\includegraphics[width=\linewidth]{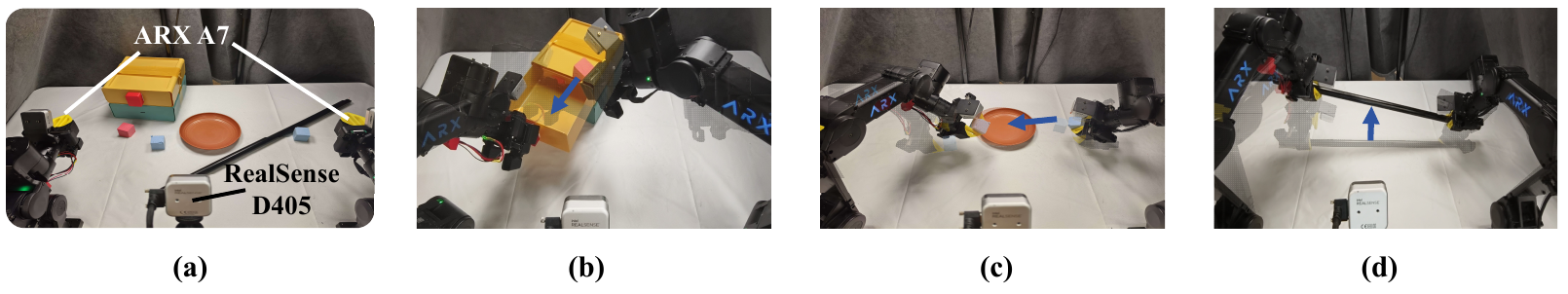}
\caption{\textbf{Real-world setup and three dual-arm tasks.}
\textbf{(a)} ARX A7 platform with a RealSense D405 camera.
\textbf{(b)} Drawer Cube Place.
\textbf{(c)} Pick Two Cube.
\textbf{(d)} Grab Stick.
}
\label{fig:real_robot}
\end{figure}

\begin{table}[!htbp]
\centering
\begin{minipage}{.70\linewidth}
\centering
\caption{\textbf{Real-robot success (\%).}
Each recipe uses one multitask StarVLA policy; 50 evaluation trials
per task. Mean weights tasks equally across all three.}
\label{tab:real_robot}
\papertablestyle
\begin{fitwidth}
\begin{tabular*}{\linewidth}{@{\extracolsep{\fill}}lrrrr@{}}
\toprule
\hiveblue{Task} & \hiveblue{V} & \hiveblue{X} & \hiveblue{X--V} & \hiveblue{\method{}}\\
\midrule
Drawer Cube Place & 24 & 34 & 38 & \rankfirst{52}\\
Pick Two Cube & 32 & 34 & 34 & \rankfirst{40}\\
Grab Stick & 38 & 46 & 76 & \rankfirst{84}\\
\midrule
\textbf{Mean} & 31.33 & 38.00 & 49.33 & \rankfirst{58.67}\\
\bottomrule
\end{tabular*}
\end{fitwidth}
\end{minipage}
\end{table}


\begin{itemize}[leftmargin=*,nosep]
    \item \textbf{Drawer Cube Place:}
    The robot opens a drawer and places a cube inside,
    requiring successful execution of successive manipulation
    stages (\cref{fig:real_robot}(b)).
    \item \textbf{Pick Two Cube:}
    The robot grasps two cubes,
    one with each gripper, and place both on the plate,
    testing coordinated control of both arms
    (\cref{fig:real_robot}(c)).
    \item \textbf{Grab Stick:}
    The robot grasps a slender rod, testing the precision
    of gripper positioning and alignment
    (\cref{fig:real_robot}(d)).
\end{itemize}

We collect 50 demonstrations per task and train one
multitask StarVLA policy for each prediction recipe:
V, X, X--V, and UniVEP.
All policies use the same demonstrations and are trained
for 20{,}000 updates with a batch size of 32.
We evaluate each policy for 50 trials per task and report
the success rate. All policies generate each action chunk using four Euler
integration steps, corresponding to four calls. We retain the original StarVLA controller across all recipes,
keeping the inference budget and action interface matched.

As shown in \cref{tab:real_robot}, UniVEP achieves
58.67\% mean success, outperforming the strongest
single-target baseline, X--V, by 9.3 percentage points.
Gains are consistent across all three tasks:
Drawer Cube Place improves from 38\% to 52\%,
Grab Stick from 76\% to 84\%, and Pick Two Cube
from 34\% to 40\%. The largest gain on Drawer Cube
Place demonstrates benefits for successive object interactions.
Qualitatively, UniVEP produces smoother arm motions
with fewer abrupt corrections during reaching and grasping,
consistent with velocity-based refinement near the
clean-action endpoint.

\section{Conclusion}
\label{sec:conclusion}



We revisited the prediction objective of flow-based robot policies
through controlled comparisons across architectures, benchmarks,
and inference budgets. On StarVLA and LiLa-WAM, velocity-weighted
endpoint prediction is particularly effective with few evaluations,
while velocity prediction benefits more from iterative refinement.
Noise-conditioned probes show complementary behavior: endpoint
estimates are less sensitive to noise initialization, whereas
velocity predictions produce more accurate near-clean corrections.

UniVEP combines these modes in a shared, consistently supervised
policy. Joint training strengthens endpoint prediction even when
inference uses only one or two evaluations; at larger budgets,
endpoint-to-velocity switching adds refinement without increasing
per-step inference FLOPs. The result is stronger few-step manipulation
and lower generation latency, with additional gains on three real
bimanual tasks. FastWAM's velocity-pretrained action head remains a
notable exception, showing that changing the prediction objective
can depend strongly on initialization. Learning the switching rule
from policy-specific evidence, rather than fixing it globally, is a promising direction for future work.

\label{cvpr:main-text-end}
\FloatBarrier
\bibliographystyle{univep_compact}
\begingroup
\setlength{\bibsep}{4.5pt}
\bibliography{references}
\endgroup
\clearpage
\phantomsection
\label{cvpr:supplement-start}
\makeappendixtitle
\appendix
\noindent\textbf{Guide to the appendix.}
\begin{itemize}[leftmargin=*,nosep,topsep=4pt]
 \item \textbf{\hyperref[sec:app_math]{\ref*{sec:app_math}. Prediction objectives.}} Loss definitions, ideal predictors, and FastWAM's velocity convention.
 \item \textbf{\hyperref[sec:app_univep]{\ref*{sec:app_univep}. UniVEP implementation.}} One-step errors, policy-specific training, and sampling grids.
 \item \textbf{\hyperref[sec:app_protocol]{\ref*{sec:app_protocol}. Evaluation protocols.}} Geometry, noise probes, uncertainty, and compute measurements.
 \item \textbf{\hyperref[sec:app_fastwam]{\ref*{sec:app_fastwam}. FastWAM controls.}} Action initialization, inference budgets, and loss weights.
 \item \textbf{\hyperref[sec:app_libero]{\ref*{sec:app_libero}. LIBERO results.}} Success by task suite and perturbation family.
 \item \textbf{\hyperref[sec:app_robotwin]{\ref*{sec:app_robotwin}. RoboTwin results.}} Per-task success rates and fixed-checkpoint sampling ablations.
\end{itemize}

\section{Prediction Objectives and Error Analysis}
\label{sec:app_math}
\subsection{Losses and the ideal predictor}
\label{sec:app_prediction}
Throughout, $x\in\R^{HD}$, $z_t=(1-t)\epsilon+tx$, $v=x-\epsilon$, and $\sigma=1-t$.
Squared $\ell_2$ norms are taken on the flattened chunk.
The symbol $v$ is the velocity of this linear path.
It is not the $v$-prediction parameterization of variance-preserving diffusion~\citep{salimans2022progressive}.
The three single-target losses are
\begin{equation}
 \begin{aligned}
 \mathcal L_{\mathrm V} &= \frac{1}{HD}\E\!\left[\norm{\hat v-v}_2^2\right],\\[3pt]
 \mathcal L_{\mathrm X} &= \frac{1}{HD}\E\!\left[\norm{\hat x-x}_2^2\right],\\[3pt]
 \mathcal L_{\mathrm{X\text{-}V}} &= \frac{1}{HD}\E\!\left[\frac{\norm{\hat x-x}_2^2}{\max(\sigma,\delta)^2}\right],
 \qquad \delta=0.05.
 \end{aligned}
 \label{eq:prediction_objectives}
\end{equation}
The expectation is over demonstrations, Gaussian noise, and the policy's training-time distribution.
All three predictions are conditioned on $(z_t,t,c)$.
Masking follows the host policy (\cref{sec:app_univep}).

For X--V, the predicted endpoint and the target endpoint are divided by the same number, so the residual is $(\hat x-x)/\max(\sigma,\delta)$.
When $\sigma\geq\delta$ this is mean squared error in velocity space.
Below the floor, the converted target is $(\sigma/\delta)v$ rather than $v$.
X and X--V use the same endpoint sampler at test time.
They differ only in which noise levels the loss emphasizes.

At a fixed $(z_t,t,c)$ with $t<1$, the conditional expectations are related by
\begin{equation}
 \E[v\mid z_t,t,c]
 =\frac{\E[x\mid z_t,t,c]-z_t}{1-t}.
 \label{eq:bayes_equivalence}
\end{equation}
A predictor that minimizes conditional squared error therefore recovers the same endpoint from either target.
Any positive weight that depends only on $t$ leaves this optimum unchanged.
It does change which noise levels a finite network emphasizes~\citep{choi2022p2,hang2023minsnr,gagneux2026training,esser2024scaling}.
That is the gap the main paper measures.
\Cref{fig:prediction_formulations} is the picture behind the algebra: X jumps to the clean action, and V takes a local step.

\noindent
\begin{minipage}[t]{.48\linewidth}
\vspace{0pt}
\paragraph{FastWAM's sign convention.}
StarVLA and LiLa-WAM step $t$ upward and regress $v=x-\epsilon$.
FastWAM steps a noise time $\sigma$ downward and regresses $u=\epsilon-x$.
We convert its native output only when we write equations in the shared notation,
\begin{equation}
 \hat v=-\hat u,\qquad
 \hat x_{\mathrm V}=z_t-\sigma\hat u=z_t+\sigma\hat v.
 \label{eq:native_velocity_signs}
\end{equation}
Reversing time and velocity together preserves the sampling update and leaves the squared losses unchanged.
The code keeps FastWAM's native convention.
The flip is only so that one set of equations covers all three policies.\end{minipage}\hfill
\begin{minipage}[t]{.48\linewidth}
\vspace{0pt}
 \centering
 \captionsetup{type=figure,font=footnotesize,skip=4pt}
 \includegraphics[width=\linewidth]{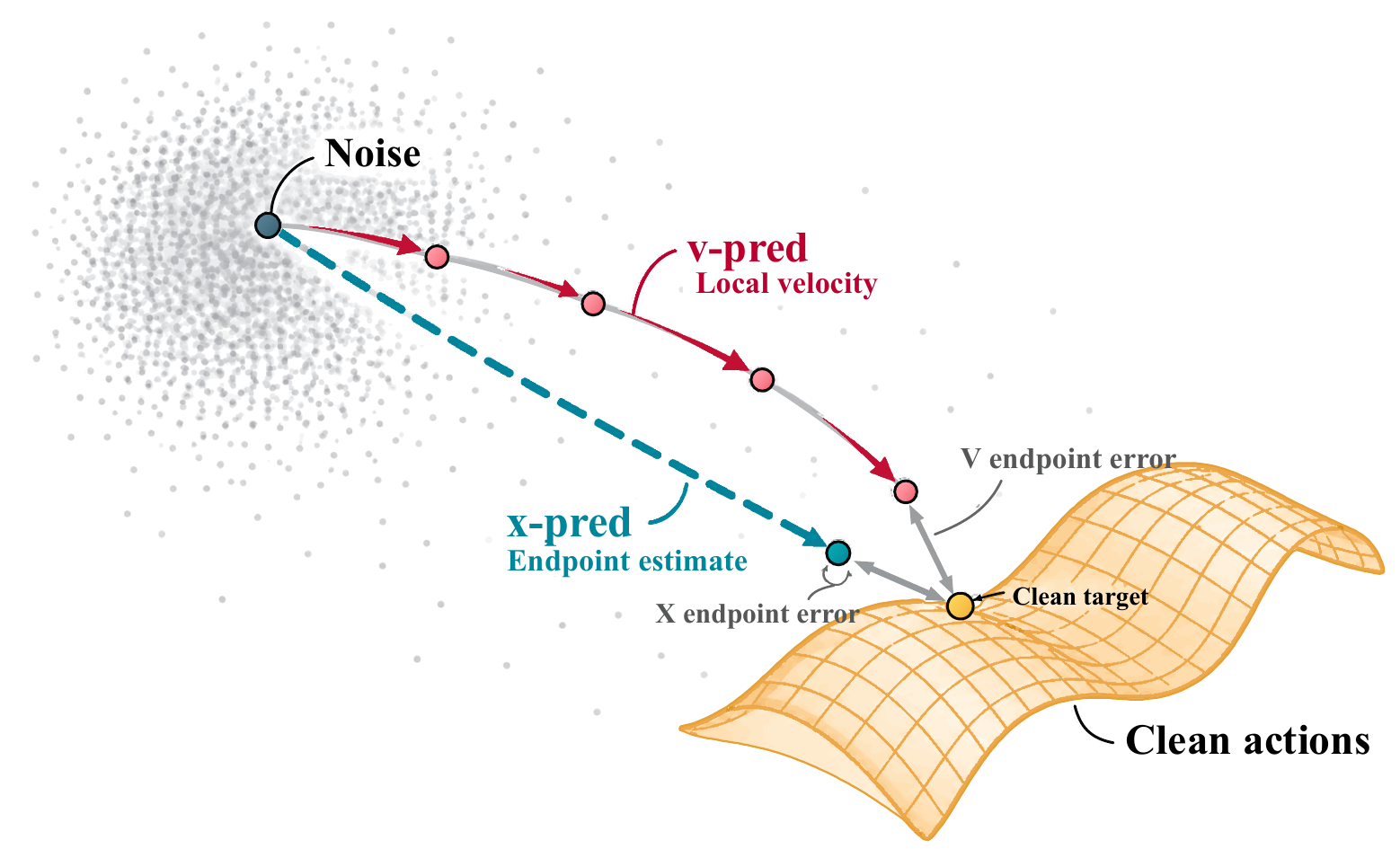}
 \caption{\textbf{Endpoint and velocity prediction on the same path.}
 X estimates the clean action, and V estimates the sample's velocity along the noise-to-action path.}
 \label{fig:prediction_formulations}
\end{minipage}\par

\section{UniVEP Implementation and Algorithms}
\label{sec:app_univep}
\subsection{Consistency and one-step error}
The consistency term in \cref{eq:ufo} sends gradients into both heads.
Applying the two updates in \cref{eq:mixed_sampling} at the same state $z_i$ gives
\begin{equation}
 z_{i+1}^{\mathrm X}-z_{i+1}^{\mathrm V}
 =\frac{h_i}{1-t_i}
  \bigl[\hat x_\theta-z_i-(1-t_i)\hat v_\theta\bigr].
 \label{eq:mode_update_agreement}
\end{equation}
Since $0<h_i\leq 1-t_i$, the mean squared difference between these updates is bounded by $\ell_{\mathrm C}$ at $z_i$.
In particular, matching the two endpoint estimates makes the two Euler updates coincide on any shared step size.
The prefactor in \cref{eq:mode_update_agreement} is the fraction of the remaining path covered by that step.

\paragraph{Error of one update.}
Let $z_{i+1}^{\rm ref}=z_i+h_i v$ be the next state on the true linear path, and write $\hat x_{\mathrm V}=z_i+\sigma_i\hat v_\theta$.
Either selected update then satisfies
\begin{equation}
 \begin{gathered}
 z_{i+1}^{m}-z_{i+1}^{\rm ref}
 =\frac{h_i}{\sigma_i}(\hat x_m-x),\\[3pt]
 \frac{\E[\norm{z_{i+1}^{m}-z_{i+1}^{\rm ref}}_2^2\mid\sigma_i]}{HD}
 =\left(\frac{h_i}{\sigma_i}\right)^2\mathcal R_m(\sigma_i).
 \end{gathered}
 \label{eq:local_update_risk}
\end{equation}
Both modes share the factor $(h_i/\sigma_i)^2$, so endpoint risk ranks their one-step errors.
This follows by expressing either update as $z_i+(h_i/\sigma_i)(\hat x_m-z_i)$ and using $x-z_i=\sigma_i v$ for the reference update. When $h_i=\sigma_i$, the step reaches the end of the grid and its error equals the endpoint error.

\subsection{How each policy is trained}
Training budgets and optimizer schedules are given in \cref{sec:training_setup,tab:recipes}.

\paragraph{Joint batch.}
StarVLA and LiLa-WAM run the two modes as two batch elements after the observation encoder.
FastWAM shares one video forward pass and uses separate attention blocks for the two action streams, with the native position indices and no attention between modes.
FastWAM averages each action loss over valid tokens, then over samples.
StarVLA averages over the unmasked chunk.
LiLa-WAM's optional per-chunk weight is off in the Full runs reported here.
Losses are computed in float32.

StarVLA draws eight noise--time pairs per demonstration and shares them across modes.

\paragraph{Losses kept from the host.}
FastWAM keeps its video loss and its action and video coefficients.
LiLa-WAM adds its future-feature loss at weight $0.5$, applied to the mean of the two modes,
\begin{equation}
 \begin{aligned}
 \mathcal L_{\mathrm{FastWAM}}
 &=\lambda_{\mathrm{video}}\mathcal L_{\mathrm{video}}
   +\lambda_{\mathrm{action}}\mathcal L_{\mathrm{UniVEP}},\\[3pt]
 \mathcal L_{\mathrm{LiLa}}
 &=\mathcal L_{\mathrm{UniVEP}}
   +\tfrac14\bigl(\mathcal L_{\mathrm{future}}^{\mathrm V}
                  +\mathcal L_{\mathrm{future}}^{\mathrm X}\bigr).
 \end{aligned}
 \label{eq:world_action_objectives}
\end{equation}
LiLa-WAM's future targets are frozen DINO features.
The three action terms use coefficient 1, and $\delta=0.05$.
Time sampling is unchanged: Beta-derived for StarVLA, shifted for FastWAM, and logit-normal for LiLa-WAM, clipped to $[0.001,0.999]$.

\paragraph{Action interface.}
StarVLA keeps its native action order and training transform, and it drops proprioception when \texttt{include\_state=false}.
Its horizon is 8 on LIBERO and 50 on RoboTwin.
FastWAM and LiLa-WAM use horizon 32.
LiLa-WAM executes 16 of those actions and keeps its native smoothing.

\subsection{Step grids and the sampler}
StarVLA and LiLa-WAM use $t_i=i/N$.
FastWAM's native shift-5 grid, written with $t=1-\sigma$, is
\begin{equation}
 t_i=\frac{i/N}{5-4i/N},\qquad i=0,\ldots,N.
 \label{eq:shift}
\end{equation}
The step allocations below use the default threshold $\sigma^*=0.5$.
The sampler tests the noise level at the start of each step. On the uniform grid, X is used for $i/N\leq 1/2$.
On FastWAM's shifted grid, solving $t_i\leq 1/2$ gives $i/N\leq 5/6$.
At 4 NFE, the uniform grid uses three X steps and one V step. On FastWAM's shifted grid, all~four~evaluations~use~endpoint~prediction.

\begin{center}
\papertablestyle
\begin{fitwidth}
\setlength{\tabcolsep}{4pt}
\begin{tabular}{@{}rcc@{}}
\toprule
\hiveblue{Calls} & \hiveblue{StarVLA / LiLa} & \hiveblue{FastWAM (shift 5)}\\
\midrule
1 & 1 X & 1 X\\
2 & 2 X & 2 X\\
4 & 3 X + 1 V & 4 X\\
8 & 5 X + 3 V & 7 X + 1 V\\
10 & 6 X + 4 V & 9 X + 1 V\\
16 & 9 X + 7 V & 14 X + 2 V\\
\bottomrule
\end{tabular}
\end{fitwidth}
\end{center}

The reversed-order controls swap X and V on the same grid.
\Cref{alg:sample} uses the shared velocity convention in \cref{eq:native_velocity_signs}.

\begin{algorithm}[!htbp]
\caption{UniVEP sampling. One head is evaluated per step.}
\label{alg:sample}
\begin{algorithmic}[1]
\REQUIRE Condition $c$, network $f_\theta$, grid $0=t_0<\cdots<t_N=1$, threshold $\sigma^*\in(0,1)$
\STATE Sample $z_0\sim\mathcal N(0,I_{HD})$
\FOR{$i=0,\ldots,N-1$}
\STATE $h_i\leftarrow t_{i+1}-t_i$; $\sigma_i\leftarrow 1-t_i$
\STATE If $\sigma_i\geq\sigma^*$, set $\hat x_\theta=f_\theta(z_i,t_i,c,1)$; else set $\hat v_\theta=f_\theta(z_i,t_i,c,0)$
\STATE Update $z_{i+1}$ with \cref{eq:mixed_sampling} and the selected prediction
\ENDFOR
\ENSURE Action chunk $z_N$, passed to the host decoder
\end{algorithmic}
\end{algorithm}

\section{Evaluation Protocols and Supplementary Evidence}
\label{sec:app_protocol}
\subsection{Action-geometry measurements}
\label{sec:app_geometry}

We sample the HIVE-Bench~\cite{liang2026hive} release of RoboTwin Clean with seed 42 using the setup in \cref{sec:geometry}.
Standardization floors the coordinate-wise standard deviation at $10^{-6}$.
Error bars in \cref{fig:geometry} report standard deviations across tasks.

\begin{figure}[!htbp]
 \centering\includegraphics[width=\linewidth]{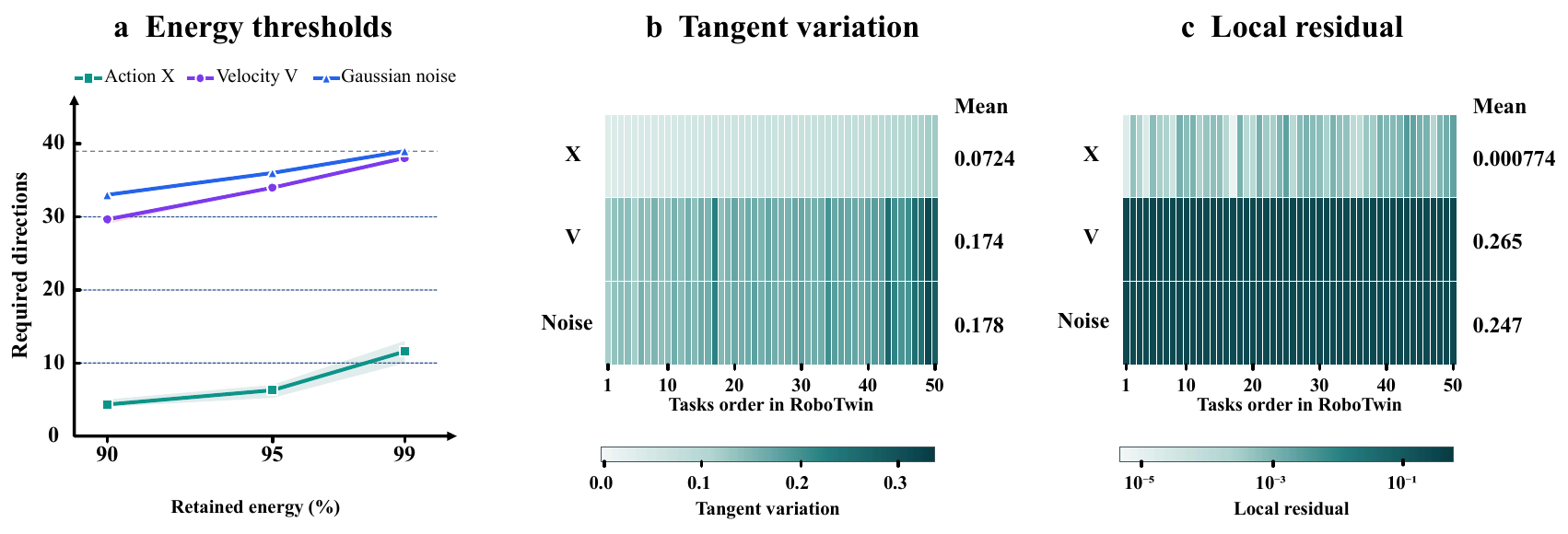}
 \caption{\textbf{Further geometry checks.}
 \textbf{(a)}~Mean rank needed to keep 90\%, 95\%, and 99\% of spectral energy.
 Bands are task-wise interquartile ranges.
 The dashed line is the rank ceiling of 39.
 \textbf{(b, c)}~Local-PCA tangent variance and mean squared residual, over 40 anchors per task.
 Rows are clean actions (X), velocity targets (V), and noise.
 Columns are the 50 tasks, ordered by increasing X tangent variance.
 Panel (b) is on a linear scale and panel (c) on a log scale.
 Numbers at the right are unweighted task means.}
 \label{fig:geometry_supplement}
\end{figure}

\paragraph{Spectral definitions.}
For each task and target we center the $40\times 448$ matrix and take its singular values $s_1\geq s_2\geq\cdots$.
Let $p_j=s_j^2/\sum_k s_k^2$.
The three spectral summaries are
\begin{equation}
 \begin{aligned}
 r_{\mathrm{stable}}&=\frac1{p_1},\\
 \mathrm{PR}&=\frac1{\sum_j p_j^2},\\
 r_\alpha&=\min\!\left\{k:\sum_{j\leq k}p_j\geq\alpha\right\}.
 \end{aligned}
 \label{eq:spectral}
\end{equation}
Mean $r_{0.90}/r_{0.95}$ values are 4.32/6.28 for clean actions, 29.66/34.00 for velocity targets, and 33.00/36.00 for noise.
Stable rank measures the total energy relative to the leading direction.
Participation ratio summarizes how evenly energy is shared across directions, reaching $k$ when $k$ directions carry equal energy.
Varying $\alpha$ in \cref{fig:geometry_supplement}(a) traces how quickly the leading directions accumulate energy.
The local PCA panels complement this view with tangent variance and residual error around individual anchors.

\paragraph{TwoNN.}
For each point $i$, let $r_{i,1}$ and $r_{i,2}$ be the distances to its nearest and second-nearest neighbors inside the same task and target.
The estimator is
\begin{equation}
 \widehat d_{\mathrm{TwoNN}}
 =\left[\frac1n\sum_{i=1}^{n}\log\frac{r_{i,2}}{r_{i,1}}\right]^{-1},
 \qquad n=40.
 \label{eq:twonn}
\end{equation}
Under the TwoNN model, uniform samples in dimension $d$ give $P(r_2/r_1\leq\mu)=1-\mu^{-d}$ for $\mu\geq 1$~\citep{facco2017twonn}, and \cref{eq:twonn} is the maximum-likelihood estimate.

\subsection{Noise probes and uncertainty}
\label{sec:app_probes}

In the probes of \cref{sec:noise_findings}, StarVLA uses the observation interface stored in its checkpoint, and LiLa-WAM uses three Full stage-2 checkpoints from epoch 4.
At each anchor, X and V receive the same conditioning and noise draws. For recovery, the input baseline in \cref{tab:low_noise_evidence} is the MSE between the corrupted chunk and the recorded clean action.
Each bootstrap replicate resamples tasks and episodes within tasks, using the same resampled units for the two predictors. This pairing is carried through to the denoising gains and V-minus-X MSE in \cref{tab:noise_gain_intervals}.

\begin{table}[!htb]
\centering
\begin{minipage}{.64\linewidth}
\setlength{\tabcolsep}{4pt}
\caption{Absolute endpoint MSE at $\sigma=0.01$ for the recovery probe in \cref{fig:noise}(b). All errors use standardized action coordinates.}
\label{tab:low_noise_evidence}
\papertablestyle
\begin{tabularx}{\linewidth}{@{}PRRR@{}}
\toprule
\hiveblue{Policy / split} & \hiveblue{Input} & \hiveblue{X} & \hiveblue{V}\\
\midrule
StarVLA / Clean   & 0.011861 & 0.087328 & 0.001539\\
StarVLA / Rand.   & 0.009942 & 0.518392 & 0.003323\\
FastWAM / Clean   & 0.000615 & 0.007474 & 0.000397\\
FastWAM / Rand.   & 0.013982 & 0.056800 & 0.013599\\
LiLa Full / Clean & 0.002938 & 0.006863 & 0.000607\\
LiLa Full / Rand. & 0.002673 & 0.056347 & 0.000617\\
\bottomrule
\end{tabularx}
\end{minipage}
\end{table}

\begin{table}[!htb]
\centering
\begin{minipage}{.78\linewidth}
\setlength{\tabcolsep}{6pt}
\caption{Paired low-noise results at $\sigma=0.01$ with 95\% task/episode bootstrap intervals.
Gain is $1-\mathrm{MSE}_{\mathrm{pred}}/\mathrm{MSE}_{\mathrm{input}}$.}
\label{tab:noise_gain_intervals}
\papertablestyle
\newcommand{\estimatewithinterval}[2]{\makecell[r]{$#1$\\[-1pt]\footnotesize $[#2]$}}
\begin{tabularx}{\linewidth}{@{}PRRR@{}}
\toprule
\hiveblue{Policy / split} & \hiveblue{X gain} & \hiveblue{V gain} & \hiveblue{MSE, V minus X}\\
\midrule
StarVLA / Clean & \estimatewithinterval{-6.362}{-24.713, 0.408} & \estimatewithinterval{0.870}{0.604, 0.956} & \estimatewithinterval{-0.08579}{-0.28468, -0.00704}\\
StarVLA / Rand. & \estimatewithinterval{-51.144}{-68.706, -35.460} & \estimatewithinterval{0.666}{0.413, 0.800} & \estimatewithinterval{-0.51507}{-0.66691, -0.35183}\\
FastWAM / Clean & \estimatewithinterval{-11.154}{-34.797, -4.789} & \estimatewithinterval{0.355}{0.152, 0.896} & \estimatewithinterval{-0.00708}{-0.01981, -0.00178}\\
FastWAM / Rand. & \estimatewithinterval{-3.062}{-119.439, -1.301} & \estimatewithinterval{0.027}{0.015, 0.684} & \estimatewithinterval{-0.04320}{-0.06467, -0.02543}\\
LiLa Full / Clean & \estimatewithinterval{-1.336}{-4.844, 0.481} & \estimatewithinterval{0.793}{0.751, 0.828} & \estimatewithinterval{-0.00626}{-0.01752, -0.00099}\\
LiLa Full / Rand. & \estimatewithinterval{-20.082}{-55.411, -0.525} & \estimatewithinterval{0.769}{0.699, 0.822} & \estimatewithinterval{-0.05573}{-0.14475, -0.00373}\\
\bottomrule
\end{tabularx}
\end{minipage}
\end{table}

For outputs $a_1,\ldots,a_R$ with $R=8$, a recorded action $x$, and a fixed standardization $S$,
\begin{equation}
 \begin{gathered}
 \bar a=\frac1R\sum_r a_r,\\
 V_R=\frac1{RHD}\sum_r\norm{S(a_r-\bar a)}_2^2,\\
 \frac1{RHD}\sum_r\norm{S(a_r-x)}_2^2
 =\frac1{HD}\norm{S(\bar a-x)}_2^2+V_R.
 \end{gathered}
 \label{eq:seed_variance}
\end{equation}
At $t=0$, the Gaussian input is independent of $x$ given $c$, and both ideal reconstructions reduce to $\E[x\mid c]$.
In \cref{eq:seed_variance}, the first term on the right measures the error of the mean reconstruction, and $V_R$ measures dispersion across draws.
\Cref{tab:one_step_stability} summarizes the latter and reports the paired V/X variance ratio for each policy and split.

\begin{table}[!htbp]
\centering
\begin{minipage}{.70\linewidth}
\setlength{\tabcolsep}{3pt}
\caption{Standardized endpoint variance across eight matched draws from pure noise, with one network call.}
\label{tab:one_step_stability}
\papertablestyle
\begin{tabular*}{\linewidth}{@{\extracolsep{\fill}}lrrrr@{}}
\toprule
\hiveblue{Policy / split} & \hiveblue{Var.\ X} & \hiveblue{Var.\ V} & \hiveblue{V/X} & \hiveblue{95\% CI}\\
\midrule
StarVLA / Clean   & 0.000064 & 0.007764 & 120.74 & $[95.48, 154.46]$\\
StarVLA / Rand.   & 0.000910 & 0.010477 & 11.51 & $[9.42, 14.82]$\\
FastWAM / Clean   & 0.000025 & 0.000601 & 24.35 & $[21.25, 27.86]$\\
FastWAM / Rand.   & 0.000312 & 0.001457 & 4.67 & $[3.73, 6.05]$\\
LiLa Full / Clean & 0.000397 & 0.001047 & 2.63 & $[2.43, 2.90]$\\
LiLa Full / Rand. & 0.000358 & 0.001078 & 3.01 & $[2.68, 3.40]$\\
\bottomrule
\end{tabular*}
\end{minipage}
\end{table}

\paragraph{Splitting risk into bias and variance.}
Let $e_m=\hat x_m-x$ and let $\bar e_m=\E_\epsilon[e_m\mid c,x,\sigma]$.
We \mbox{measure both terms in the same action metric.}\vphantom{,}
\begingroup
\abovedisplayskip=\abovedisplayshortskip
\belowdisplayskip=\belowdisplayshortskip
\begin{equation}
 \begin{aligned}
 B_m(\sigma)&=\frac{1}{HD}\E_{c,x}\norm{\bar e_m}_2^2,\\
 V_m(\sigma)&=\frac{1}{HD}\E_{c,x,\epsilon}\norm{e_m-\bar e_m}_2^2,\\
 \mathcal R_m(\sigma)&=B_m(\sigma)+V_m(\sigma).
 \end{aligned}
 \label{eq:risk_decomposition}
\end{equation}
\endgroup
Applying the same decomposition to $\hat v_\theta-v$ gives $B_v$ and $V_v$ and, by \cref{eq:low_noise_identity}, $\mathcal R_{\mathrm V}=\sigma^2(B_v+V_v)$.
The scaling applies to both the squared conditional bias and the conditional variance of the velocity error. The recovery probe in \cref{tab:noise_gain_intervals} measures improvement over the corrupted input directly.

\subsection{How latency and FLOPs are measured}
\label{sec:app_compute}

We time each checkpoint in a single process from synthetic images and joint state on the CPU to the CPU action chunk, using batch size one on an RTX 4090.
After 20 warm-up calls, we collect 300 synchronized timings.
We count two FLOPs per multiply-add and report the dominant cost of matrix multiplications, convolutions, and attention operations.

On LIBERO, StarVLA uses 1.908~TFLOPs per chunk at 4 NFE, with latencies of 80.52, 82.34, 79.68, and 80.61~ms for V, X, X--V, and UniVEP, respectively.
FastWAM scratch-action UniVEP uses 3.737~TFLOPs and 337.60~ms per chunk at 10 NFE.

\subsection{Training cost}
\label{sec:app_training_cost}

\noindent
\begin{minipage}[t]{.50\linewidth}
\vspace{0pt}
We time StarVLA on one A800 with 16 LIBERO demonstrations per update and eight noise repeats.
All recipes train the full model with the same precision, eager vision--language attention, and gradient checkpointing.
We average ten post-warm-up updates from the forward pass through backpropagation, gradient clipping, and AdamW~\cite{loshchilov2017decoupled}.
Counted FLOPs cover the dominant forward and backward operations, including checkpoint recomputation.
\end{minipage}\hfill
\begin{minipage}[t]{.46\linewidth}
\vspace{0pt}
\centering
\captionsetup{type=table}
\caption{StarVLA training cost. Time is mean $\pm$ standard deviation. GiB reports peak allocated device memory.}
\label{tab:training_cost}
\papertablestyle
\begin{tabularx}{\linewidth}{@{}PRRR@{}}
\toprule
\hiveblue{Recipe} & \hiveblue{TFLOPs} & \hiveblue{Seconds} & \hiveblue{GiB}\\
\midrule
V & 114.47 & $2.110\pm0.061$ & 44.65\\
X & 114.47 & $2.095\pm0.053$ & 44.65\\
X--V & 114.47 & $2.098\pm0.050$ & 44.65\\
\tmethod
\textbf{\method{}} & 122.40 & $2.439\pm0.045$ & 47.44\\
\bottomrule
\end{tabularx}
\end{minipage}\par

\Needspace{10\baselineskip}
\section{FastWAM Initialization and Additional Controls}
\label{sec:app_fastwam}

\subsection{Action initialization}
\label{sec:app_fastwam_initialization}

The paired runs change the initialization of the action expert while retaining the pretrained video backbone and the joint video--action objective. X and X--V predict the same clean action and use the same endpoint update. Their different gains from scratch initialization (3.32 and 30.28~pp) locate the largest sensitivity in the recipe that gives extra weight to low-noise endpoint errors.

\noindent\makebox[\linewidth][c]{%
\begin{minipage}{.64\linewidth}
\captionsetup{type=table}
\caption{FastWAM on RoboTwin Clean at 10 NFE. Scratch resets ActionDiT and retains the pretrained video backbone. $\Delta$ is scratch minus pretrained (pp).}
\label{tab:initialization}
\papertablestyle
\begin{tabularx}{\linewidth}{@{}PRRR@{}}
\toprule
\hiveblue{Target} & \hiveblue{Pretrained} & \hiveblue{Scratch} & \hiveblue{$\boldsymbol{\Delta}$}\\
\midrule
V & \textbf{73.24} & \textbf{71.56} & $-1.68$\\
X & 65.88 & 69.20 & $+3.32$\\
X--V & 34.20 & 64.48 & $+30.28$\\
\bottomrule
\end{tabularx}
\end{minipage}%
}\par

\subsection{Inference-budget sweep}
With pretrained actions, X--V stays near 35\% across budgets. With scratch actions, it reaches 59.64--65.12\%.
The sweep uses separate rollouts from the native-budget comparison in \cref{tab:initialization}.
Per-task scores are in \cref{tab:app_fast_nfe,tab:app_fast_ufo}.
The ranking also changes with inference budget in \cref{tab:app_fast_nfe_summary}. With pretrained actions, UniVEP--X reaches 66.08\% with one evaluation, and V has the highest success at every tested budget from 2 to 16 NFE. \Cref{sec:app_univep} lists the number of X and V updates that UniVEP--X uses on FastWAM's~shifted~grid~at~each~inference~budget.

\begin{table}[!htbp]
\centering
\begin{minipage}{.76\linewidth}
\setlength{\tabcolsep}{2.5pt}
\caption{FastWAM success (\%) on RoboTwin Clean across the call budget.
Green marks the best success rate at each budget within an initialization block.}
\label{tab:app_fast_nfe_summary}
\papertablestyle
\begin{tabularx}{\linewidth}{@{}PRRRRRR@{}}
\toprule
\hiveblue{Recipe} & \hiveblue{1} & \hiveblue{2} & \hiveblue{4} & \hiveblue{8} & \hiveblue{10} & \hiveblue{16}\\
\midrule
\tgroup
\multicolumn{7}{@{}P}{\textbf{(a) Pretrained action head}}\\
V & 57.20 & \rankfirst{69.64} & \rankfirst{71.24} & \rankfirst{73.36} & \rankfirst{72.64} & \rankfirst{72.52}\\
X & 59.64 & 61.76 & 62.80 & 65.64 & 65.72 & 66.92\\
X--V & 32.64 & 34.24 & 35.64 & 35.00 & 34.96 & 35.48\\
\tmethod
\textbf{\method{}} & 59.60 & 59.80 & 61.24 & 57.56 & 57.00 & 57.68\\
\method{}--X & \rankfirst{66.08} & 66.16 & 66.04 & 66.28 & 66.36 & 66.32\\
\midrule
\tgroup
\multicolumn{7}{@{}P}{\textbf{(b) Scratch action head}}\\
V & 43.24 & \rankfirst{68.20} & \rankfirst{70.04} & \rankfirst{71.32} & \rankfirst{70.84} & \rankfirst{73.12}\\
X & \rankfirst{65.68} & 66.52 & 69.50 & 67.48 & 68.20 & 68.96\\
X--V & 60.84 & 59.64 & 62.20 & 63.12 & 64.40 & 65.12\\
\tmethod
\textbf{\method{}} & 55.92 & 56.24 & 57.24 & 57.04 & 56.12 & 56.68\\
\bottomrule
\end{tabularx}
\end{minipage}
\end{table}

\FloatBarrier
\subsection{Loss-weight ablation}
In FastWAM UniVEP--X, changing the coefficients of $(\ell_{\mathrm V},\ell_{\mathrm X},\ell_{\mathrm C})$ from $(1,1,1)$ to $(1,0.2,0.1)$ raises RoboTwin Clean success from 66.36\% to 68.18\% at 10 NFE.
This increases the relative weight of V supervision within the action loss. The 1.82~pp gain reflects the combined reduction in the weights on \mbox{the endpoint and consistency losses.}

\subsection{Additional inference costs}
\Cref{tab:app_fast_compute} reports the scratch-action RoboTwin and pretrained LIBERO measurements using the protocol in \cref{sec:app_compute}.
\Cref{tab:compute_nfe_latency} gives the RoboTwin sweep for pretrained actions.

\begin{table}[!htbp]
\centering
\begin{minipage}{.66\linewidth}
\caption{FastWAM inference cost at 10 NFE. FLOPs are TFLOPs per chunk, and recipe rows report latency in ms per chunk.}
\label{tab:app_fast_compute}
\papertablestyle
\begin{tabularx}{\linewidth}{@{}PRR@{}}
\toprule
\hiveblue{Metric / recipe} & \hiveblue{RoboTwin (scratch)} & \hiveblue{LIBERO (pretrained)}\\
\midrule
TFLOPs & 4.041 & 3.737\\
\midrule
V & 364.49 & 336.98\\
X & 369.10 & 334.05\\
X--V & 369.40 & 338.24\\
\method{} & 367.90 & 328.20\\
\method{}--X & -- & 334.54\\
\bottomrule
\end{tabularx}
\end{minipage}
\end{table}

\FloatBarrier

\newlength{\appresulttaskwidth}
\newlength{\appresultnumberwidth}
\newcolumntype{T}{>{\raggedright\arraybackslash}p{\appresulttaskwidth}}
\newcolumntype{N}{>{\raggedleft\arraybackslash}p{\appresultnumberwidth}}
\newenvironment{appendixresultstable}[2][.84]{%
  \par\begingroup\centering
  \begin{minipage}{#1\linewidth}
  \centering\papertablestyle
  \setlength{\tabcolsep}{0.75pt}%
  \setlength{\appresulttaskwidth}{104pt}%
  \setlength{\appresultnumberwidth}{\dimexpr(\linewidth-\appresulttaskwidth)/#2-2\tabcolsep\relax}%
  \captionsetup{type=table,width=\linewidth,justification=raggedright,singlelinecheck=false}%
}{\end{minipage}\par\endgroup}

\Needspace{22\baselineskip}
\section{LIBERO Suites and Perturbations}
\label{sec:app_libero}
Green marks the best value in each column within a policy block.
The same LIBERO-trained checkpoints are evaluated on the original suites and on the perturbed scenes.
For StarVLA, X--V leads in average LIBERO success, and UniVEP leads in overall LIBERO-Plus success.
For LiLa-WAM, UniVEP leads on both aggregates and has the highest scores on the camera, lighting, background, and layout perturbations.

\begin{table}[!htb]
\centering
\begin{minipage}{.75\linewidth}
\setlength{\tabcolsep}{3pt}
\caption{LIBERO success (\%).
AVG is the unweighted mean of Spatial, Object, Goal, and Long.}
\label{tab:app_libero}
\papertablestyle
\begin{tabularx}{\linewidth}{@{}PRRRRR@{}}
\toprule
\hiveblue{Recipe} & \hiveblue{Spa.} & \hiveblue{Obj.} & \hiveblue{Goal} & \hiveblue{Long} & \hiveblue{AVG}\\
\midrule
\tgroup
\multicolumn{6}{@{}P}{\textbf{StarVLA}}\\
V & \arxivthird{98.80} & \arxivsecond{99.20} & 98.20 & 93.60 & 97.45\\
X & \arxivfirst{\textbf{99.40}} & \arxivfirst{\textbf{99.40}} & \arxivthird{98.40} & \arxivsecond{95.40} & \arxivsecond{98.15} \\
X--V & \arxivsecond{99.00} & \arxivfirst{\textbf{99.40}} & \arxivsecond{98.80} & \arxivfirst{\textbf{96.60}} & \arxivfirst{\textbf{98.45}} \\
\tmethod
\textbf{\method{}} & 98.40 & \arxivthird{98.60} & \arxivfirst{\textbf{99.60}} & \arxivthird{94.40} & \arxivthird{97.75} \\
\midrule
\tgroup
\multicolumn{6}{@{}P}{\textbf{FastWAM}}\\
V & \arxivthird{97.00} & \arxivthird{99.40} & 96.40 & 92.00 & 96.20\\
X & 96.60 & 98.40 & 96.40 & 90.40 & 95.45\\
X--V & \arxivsecond{97.80} & \arxivfirst{\textbf{100}} & 97.20 & \arxivthird{92.20} & \arxivthird{96.80} \\
\tmethod
\textbf{\method{}} & 96.50 & \arxivsecond{99.50} & \arxivthird{98.00} & \arxivsecond{95.00} & \arxivsecond{97.25} \\
\method{}, scratch & \arxivfirst{\textbf{98.50}} & 98.50 & \arxivfirst{\textbf{100}} & \arxivfirst{\textbf{96.50}} & \arxivfirst{\textbf{98.38}} \\
\method{}--X & \arxivthird{97.00} & 97.50 & \arxivsecond{99.50} & 91.00 & 96.25\\
\midrule
\tgroup
\multicolumn{6}{@{}P}{\textbf{LiLa-WAM}}\\
V & \arxivthird{97.20} & \arxivfirst{\textbf{98.40}} & \arxivthird{98.20} & 90.60 & \arxivthird{96.10} \\
X & 96.80 & \arxivfirst{\textbf{98.40}} & 97.20 & \arxivthird{91.60} & 96.00\\
X--V & \arxivsecond{97.80} & \arxivthird{97.60} & \arxivsecond{98.40} & \arxivsecond{92.80} & \arxivsecond{96.65} \\
\tmethod
\textbf{\method{}} & \arxivfirst{\textbf{98.20}} & \arxivsecond{98.00} & \arxivfirst{\textbf{98.80}} & \arxivfirst{\textbf{93.40}} & \arxivfirst{\textbf{97.10}} \\
\bottomrule
\end{tabularx}
\end{minipage}
\end{table}

\begin{table}[!htb]
\centering
\begin{minipage}{.86\linewidth}
\caption{LIBERO-Plus success rates (\%) grouped by perturbation family for each policy.
The Total column reports the \mbox{overall success rate across the perturbed benchmark.}}
\label{tab:app_plus}
\papertablestyle
\begin{tabularx}{\linewidth}{@{}PRRRRRRRR@{}}
\toprule
\hiveblue{Recipe} & \hiveblue{Cam.} & \hiveblue{Robot} & \hiveblue{Lang.} & \hiveblue{Light} & \hiveblue{Backgr.} & \hiveblue{Noise} & \hiveblue{Layout} & \hiveblue{Total}\\
\midrule
\tgroup
\multicolumn{9}{@{}P}{\textbf{StarVLA}}\\
V & 47.50 & 57.80 & 86.70 & \arxivsecond{96.00} & \arxivsecond{96.70} & 78.20 & 79.50 & 75.70\\
X & \arxivthird{51.20} & \arxivsecond{65.20} & \arxivsecond{87.50} & \arxivfirst{\textbf{97.20}} & 95.50 & \arxivthird{80.80} & \arxivthird{81.00} & \arxivthird{78.20} \\
X--V & \arxivsecond{55.00} & \arxivfirst{\textbf{65.40}} & \arxivthird{87.40} & 94.90 & \arxivfirst{\textbf{97.10}} & \arxivfirst{\textbf{84.40}} & \arxivfirst{\textbf{83.10}} & \arxivsecond{79.60} \\
\tmethod
\textbf{\method{}} & \arxivfirst{\textbf{59.80}} & \arxivthird{63.50} & \arxivfirst{\textbf{91.70}} & \arxivthird{95.60} & \arxivthird{95.60} & \arxivsecond{84.10} & \arxivsecond{82.00} & \arxivfirst{\textbf{80.40}} \\
\midrule
\tgroup
\multicolumn{9}{@{}P}{\textbf{FastWAM}}\\
V & 15.60 & 41.90 & \arxivthird{68.30} & \arxivthird{81.70} & \arxivsecond{52.50} & 42.90 & 60.30 & 50.40\\
X & 13.60 & \arxivfirst{\textbf{50.90}} & 65.70 & 78.50 & 51.00 & 41.20 & \arxivthird{60.90} & 50.30\\
X--V & \arxivthird{24.60} & 31.10 & 59.30 & 77.80 & 48.80 & 41.70 & 57.60 & 47.30\\
\tmethod
\textbf{\method{}} & \arxivsecond{33.80} & 33.30 & 67.10 & 80.30 & \arxivthird{51.60} & \arxivthird{48.10} & 60.50 & \arxivthird{52.40} \\
\method{}, scratch & \arxivfirst{\textbf{54.90}} & \arxivthird{44.30} & \arxivfirst{\textbf{80.80}} & \arxivsecond{86.10} & 50.00 & \arxivfirst{\textbf{73.50}} & \arxivfirst{\textbf{72.90}} & \arxivfirst{\textbf{66.00}} \\
\method{}--X & 8.90 & \arxivsecond{46.60} & \arxivsecond{70.30} & \arxivfirst{\textbf{88.40}} & \arxivfirst{\textbf{65.90}} & \arxivsecond{50.80} & \arxivsecond{66.30} & \arxivsecond{54.70} \\
\midrule
\tgroup
\multicolumn{9}{@{}P}{\textbf{LiLa-WAM}}\\
V & 21.50 & 48.40 & \arxivfirst{\textbf{69.10}} & \arxivthird{68.80} & 51.20 & 49.70 & 57.40 & 52.30\\
X & \arxivsecond{23.40} & \arxivthird{53.30} & 65.70 & 65.10 & \arxivthird{53.50} & \arxivfirst{\textbf{54.80}} & \arxivsecond{62.20} & \arxivthird{54.00} \\
X--V & \arxivthird{21.60} & \arxivsecond{58.20} & \arxivthird{68.40} & \arxivsecond{72.10} & \arxivsecond{56.30} & \arxivthird{50.80} & \arxivthird{60.40} & \arxivsecond{55.40} \\
\tmethod
\textbf{\method{}} & \arxivfirst{\textbf{26.30}} & \arxivfirst{\textbf{60.20}} & \arxivsecond{68.80} & \arxivfirst{\textbf{74.30}} & \arxivfirst{\textbf{56.60}} & \arxivsecond{50.90} & \arxivfirst{\textbf{71.70}} & \arxivfirst{\textbf{58.40}} \\
\bottomrule
\end{tabularx}
\end{minipage}
\end{table}
\FloatBarrier

\Needspace{36\baselineskip}
\section{Per-Task RoboTwin Results}
\label{sec:app_robotwin}
The AVG rows weight all 50 RoboTwin tasks equally. C and R denote Clean and Randomized.
The single-target sweeps show how individual tasks respond to added evaluations. The sampling-order tables in \cref{tab:app_route_ablation,tab:app_lila_reverse} compare schedules using fixed trained checkpoints. At 4 NFE, StarVLA scores 48\% with X$\to$V and 14\% with V$\to$X on Open Microwave, and 42\% and 80\% on Handover Mic.

\begin{figure}[!htb]
 \centering\includegraphics[width=0.98\textwidth]{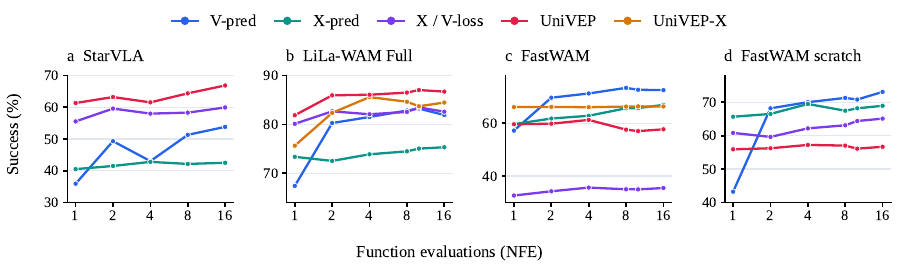}
 \caption{\textbf{RoboTwin Clean success by inference budget.}
 StarVLA and LiLa-WAM Full use the three-seed means in \cref{tab:nfe}; FastWAM uses the sweep in \cref{tab:app_fast_nfe_summary}.}
 \label{fig:nfe}
\end{figure}

\begin{appendixresultstable}[.88]{6}
\captionsetup{width=\linewidth,skip=3pt}
\caption{LiLa-WAM Full, single-target Clean success (\%) at 10 calls.}\label{tab:app_lila_base}
\centering
\begin{tabularx}{\linewidth}{@{}PRRR@{\hspace{12pt}}PRRR@{}}
\toprule
\hiveblue{Task} & \hiveblue{$v$} & \hiveblue{$x$} & \hiveblue{$x$--V} & \hiveblue{Task} & \hiveblue{$v$} & \hiveblue{$x$} & \hiveblue{$x$--V}\\
\midrule
Adjust Bottle & 100 & 100 & 100 & Place Can Basket & 78 & 58 & 84\\
Beat Block Hammer & 98 & 92 & 92 & Place Cans Plasticbox & 100 & 94 & 98\\
Blocks Ranking RGB & 100 & 98 & 100 & Place Container Plate & 100 & 98 & 98\\
Blocks Ranking Size & 84 & 78 & 84 & Place Dual Shoes & 46 & 50 & 46\\
Click Alarmclock & 84 & 62 & 90 & Place Empty Cup & 80 & 78 & 82\\
Click Bell & 66 & 38 & 72 & Place Fan & 92 & 68 & 90\\
Dump Bin Bigbin & 92 & 94 & 86 & Place Mouse Pad & 42 & 50 & 56\\
\bottomrule
\end{tabularx}
\end{appendixresultstable}

\noindent\makebox[\linewidth][c]{%
\begin{minipage}[t]{.50\linewidth}
\vspace{0pt}
\begin{appendixresultstable}[1]{6}
\setlength{\appresulttaskwidth}{99pt}
\setlength{\appresultnumberwidth}{\dimexpr(\linewidth-\appresulttaskwidth)/6-2\tabcolsep\relax}

\captionsetup{width=\linewidth,skip=3pt}

\caption{StarVLA single-target success (\%) at 4 NFE.}\label{tab:app_star_base}

\papertablelook
\begin{fitwidth}
\begin{tabular}{@{}T*{6}{N}@{}}
\toprule
\hiveblue{Task} & \multicolumn{2}{c}{\hiveblue{$v$}} & \multicolumn{2}{c}{\hiveblue{$x$}} & \multicolumn{2}{c}{\hiveblue{$x$--V}}\\
\cmidrule(lr){2-3} \cmidrule(lr){4-5} \cmidrule(lr){6-7}
 & \hiveblue{C} & \hiveblue{R} & \hiveblue{C} & \hiveblue{R} & \hiveblue{C} & \hiveblue{R}\\
\midrule
Adjust Bottle & 98 & 0 & 96 & 2 & 100 & 6\\
Beat Block Hammer & 36 & 0 & 78 & 8 & 82 & 16\\
Blocks Ranking RGB & 24 & 0 & 24 & 0 & 74 & 2\\
Blocks Ranking Size & 0 & 0 & 2 & 4 & 24 & 4\\
Click Alarmclock & 12 & 54 & 14 & 22 & 10 & 28\\
Click Bell & 52 & 18 & 8 & 8 & 2 & 4\\
\bottomrule
\end{tabular}
\end{fitwidth}
\end{appendixresultstable}

\end{minipage}\hspace{.02\linewidth}%
\begin{minipage}[t]{.45\linewidth}
\vspace{0pt}
\begin{appendixresultstable}[1]{5}
\setlength{\appresulttaskwidth}{99pt}
\setlength{\appresultnumberwidth}{\dimexpr(\linewidth-\appresulttaskwidth)/5-2\tabcolsep\relax}

\captionsetup{width=\linewidth,skip=3pt}

\caption{StarVLA UniVEP on Clean, by number of calls.}\label{tab:app_star_ufo}

\papertablelook
\begin{fitwidth}
\begin{tabular}{@{}T*{5}{N}@{}}
\toprule
\hiveblue{Task} & \multicolumn{5}{c}{\hiveblue{NFE}}\\
\cmidrule(lr){2-6}
 & \hiveblue{1} & \hiveblue{2} & \hiveblue{4} & \hiveblue{8} & \hiveblue{16}\\
\midrule
Adjust Bottle & 100 & 100 & 100 & 100 & 100\\
Beat Block Hammer & 90 & 90 & 94 & 96 & 94\\
Blocks Ranking RGB & 58 & 78 & 70 & 78 & 78\\
Blocks Ranking Size & 24 & 30 & 30 & 32 & 32\\
Click Alarmclock & 0 & 0 & 0 & 0 & 0\\
Click Bell & 2 & 0 & 2 & 0 & 0\\
\bottomrule
\end{tabular}
\end{fitwidth}
\end{appendixresultstable}
\end{minipage}%
}\par

\clearpage
\begin{center}
\begin{minipage}{.77\linewidth}
\captionof{table}{Per-task RoboTwin Clean success (\%) with the consistency loss removed.
StarVLA uses UniVEP at 4 NFE, and FastWAM uses UniVEP--X at 10 NFE.}\label{tab:app_consistency_ablation}
\papertablestyle
\begin{tabularx}{\linewidth}{@{}PRR@{}}
\toprule
\hiveblue{Task} & \hiveblue{StarVLA} & \hiveblue{FastWAM}\\
\midrule
Adjust Bottle & 100 & 98\\
Beat Block Hammer & 74 & 68\\
Blocks Ranking RGB & 52 & 26\\
Blocks Ranking Size & 28 & 12\\
Click Alarmclock & 0 & 92\\
Click Bell & 0 & 94\\
Dump Bin Bigbin & 94 & 96\\
Grab Roller & 100 & 100\\
Handover Block & 20 & 10\\
Handover Mic & 38 & 60\\
Hanging Mug & 16 & 8\\
Lift Pot & 86 & 82\\
Move Can Pot & 42 & 86\\
Move Pillbottle Pad & 74 & 74\\
Move Playingcard Away & 62 & 70\\
Move Stapler Pad & 20 & 4\\
Open Laptop & 96 & 72\\
Open Microwave & 26 & 50\\
Pick Diverse Bottles & 40 & 44\\
Pick Dual Bottles & 62 & 72\\
Place A2B Left & 16 & 44\\
Place A2B Right & 28 & 60\\
Place Bread Basket & 60 & 70\\
Place Bread Skillet & 72 & 74\\
Place Burger Fries & 96 & 94\\
Place Can Basket & 32 & 52\\
Place Cans Plasticbox & 98 & 96\\
Place Container Plate & 96 & 90\\
Place Dual Shoes & 60 & 26\\
Place Empty Cup & 80 & 94\\
Place Fan & 58 & 40\\
Place Mouse Pad & 32 & 14\\
Place Object Basket & 64 & 14\\
Place Object Scale & 46 & 48\\
Place Object Stand & 68 & 88\\
Place Phone Stand & 36 & 48\\
Place Shoe & 68 & 60\\
Press Stapler & 92 & 82\\
Put Bottles Dustbin & 36 & 4\\
Put Object Cabinet & 34 & 42\\
Rotate QRcode & 52 & 46\\
Scan Object & 72 & 68\\
Shake Bottle Horizontally & 98 & 100\\
Shake Bottle & 96 & 100\\
Stack Blocks Three & 32 & 20\\
Stack Blocks Two & 76 & 64\\
Stack Bowls Three & 64 & 50\\
Stack Bowls Two & 74 & 80\\
Stamp Seal & 60 & 34\\
Turn Switch & 50 & 44\\
\midrule
\textbf{AVG} & \textbf{57.52} & \textbf{59.28}\\
\bottomrule
\end{tabularx}
\end{minipage}
\end{center}

\clearpage
\begin{appendixresultstable}[.95]{15}
\caption{StarVLA single-target NFE sweep, Clean success rates (\%).}\label{tab:app_star_nfe}
\papertablelook
\begin{fitwidth}
\begin{tabular}{@{}T*{15}{N}@{}}
\toprule
\hiveblue{Task} & \multicolumn{3}{c}{\hiveblue{NFE=1}} & \multicolumn{3}{c}{\hiveblue{NFE=2}} & \multicolumn{3}{c}{\hiveblue{NFE=4}} & \multicolumn{3}{c}{\hiveblue{NFE=8}} & \multicolumn{3}{c}{\hiveblue{NFE=16}}\\
\cmidrule(lr){2-4} \cmidrule(lr){5-7} \cmidrule(lr){8-10} \cmidrule(lr){11-13} \cmidrule(lr){14-16}
 & \hiveblue{$v$} & \hiveblue{$x$} & \hiveblue{$x$--V} & \hiveblue{$v$} & \hiveblue{$x$} & \hiveblue{$x$--V} & \hiveblue{$v$} & \hiveblue{$x$} & \hiveblue{$x$--V} & \hiveblue{$v$} & \hiveblue{$x$} & \hiveblue{$x$--V} & \hiveblue{$v$} & \hiveblue{$x$} & \hiveblue{$x$--V}\\
\midrule
Adjust Bottle & 96 & 98 & 100 & 100 & 96 & 100 & 100 & 100 & 100 & 100 & 96 & 100 & 98 & 100 & 100\\
Beat Block Hammer & 32 & 64 & 72 & 64 & 62 & 80 & 34 & 60 & 92 & 74 & 68 & 84 & 86 & 62 & 84\\
Blocks Ranking RGB & 22 & 18 & 62 & 54 & 28 & 74 & 20 & 18 & 64 & 40 & 16 & 68 & 60 & 22 & 70\\
Blocks Ranking Size & 4 & 4 & 20 & 28 & 6 & 18 & 6 & 8 & 26 & 4 & 4 & 24 & 32 & 10 & 36\\
Click Alarmclock & 6 & 6 & 0 & 4 & 8 & 0 & 8 & 12 & 0 & 6 & 12 & 0 & 2 & 10 & 0\\
Click Bell & 8 & 6 & 4 & 2 & 8 & 4 & 54 & 10 & 0 & 2 & 6 & 0 & 4 & 2 & 0\\
\bottomrule
\end{tabular}
\end{fitwidth}
\end{appendixresultstable}

\clearpage
\begin{appendixresultstable}{12}
\caption{FastWAM single-target success rates (\%): pretrained and scratch action initializations.}\label{tab:app_fast_base}
\papertablelook
\begin{fitwidth}
\begin{tabular}{@{}T*{12}{N}@{}}
\toprule
\hiveblue{Task} & \multicolumn{6}{c}{\hiveblue{Pretrained action}} & \multicolumn{6}{c}{\hiveblue{Scratch action}}\\
 & \multicolumn{2}{c}{\hiveblue{$v$}} & \multicolumn{2}{c}{\hiveblue{$x$}} & \multicolumn{2}{c}{\hiveblue{$x$--V}} & \multicolumn{2}{c}{\hiveblue{$v$}} & \multicolumn{2}{c}{\hiveblue{$x$}} & \multicolumn{2}{c}{\hiveblue{$x$--V}}\\
 & \hiveblue{C} & \hiveblue{R} & \hiveblue{C} & \hiveblue{R} & \hiveblue{C} & \hiveblue{R} & \hiveblue{C} & \hiveblue{R} & \hiveblue{C} & \hiveblue{R} & \hiveblue{C} & \hiveblue{R}\\
\midrule
Adjust Bottle & 98 & 0 & 96 & 0 & 76 & 0 & 100 & 0 & 100 & 0 & 100 & 0\\
Beat Block Hammer & 88 & 0 & 80 & 0 & 62 & 0 & 100 & 0 & 94 & 0 & 88 & 0\\
Blocks Ranking RGB & 78 & 0 & 66 & 0 & 14 & 0 & 72 & 0 & 70 & 0 & 88 & 0\\
Blocks Ranking Size & 42 & 0 & 32 & 0 & 0 & 0 & 28 & 0 & 32 & 0 & 50 & 0\\
Click Alarmclock & 98 & 28 & 86 & 30 & 62 & 6 & 98 & 64 & 98 & 52 & 90 & 36\\
\bottomrule
\end{tabular}
\end{fitwidth}
\end{appendixresultstable}

\clearpage
\begin{appendixresultstable}{12}
\caption{FastWAM single-target NFE sweep, Clean success rates (\%): pretrained and scratch action heads, NFE=1, 2}\label{tab:app_fast_nfe}\label{tab:app_fast_scratch}
\papertablelook
\begin{fitwidth}
\begin{tabular}{@{}T*{12}{N}@{}}
\toprule
\hiveblue{Task} & \multicolumn{6}{c}{\hiveblue{NFE=1}} & \multicolumn{6}{c}{\hiveblue{NFE=2}}\\
\cmidrule(lr){2-7} \cmidrule(lr){8-13}
 & \multicolumn{3}{c}{\hiveblue{Pretrained}} & \multicolumn{3}{c}{\hiveblue{Scratch}} & \multicolumn{3}{c}{\hiveblue{Pretrained}} & \multicolumn{3}{c}{\hiveblue{Scratch}}\\
 & \hiveblue{$v$} & \hiveblue{$x$} & \hiveblue{$x$--V} & \hiveblue{$v$} & \hiveblue{$x$} & \hiveblue{$x$--V} & \hiveblue{$v$} & \hiveblue{$x$} & \hiveblue{$x$--V} & \hiveblue{$v$} & \hiveblue{$x$} & \hiveblue{$x$--V}\\
\midrule
Adjust Bottle & 96 & 78 & 88 & 96 & 100 & 100 & 92 & 86 & 88 & 100 & 100 & 92\\
Beat Block Hammer & 78 & 42 & 38 & 50 & 92 & 80 & 92 & 64 & 62 & 96 & 96 & 82\\
Blocks Ranking RGB & 42 & 64 & 12 & 10 & 66 & 76 & 64 & 74 & 10 & 54 & 60 & 74\\
Blocks Ranking Size & 12 & 36 & 2 & 6 & 26 & 40 & 32 & 30 & 0 & 24 & 32 & 50\\
Click Alarmclock & 100 & 82 & 72 & 100 & 98 & 84 & 96 & 86 & 70 & 100 & 98 & 84\\
\bottomrule
\end{tabular}
\end{fitwidth}
\end{appendixresultstable}

\clearpage
\begin{appendixresultstable}{12}
\ContinuedFloat
\caption[]{FastWAM single-target NFE sweep, Clean success rates (\%): pretrained and scratch action heads, NFE=4, 8 (continued).}
\papertablelook
\begin{fitwidth}
\begin{tabular}{@{}T*{12}{N}@{}}
\toprule
\hiveblue{Task} & \multicolumn{6}{c}{\hiveblue{NFE=4}} & \multicolumn{6}{c}{\hiveblue{NFE=8}}\\
\cmidrule(lr){2-7} \cmidrule(lr){8-13}
 & \multicolumn{3}{c}{\hiveblue{Pretrained}} & \multicolumn{3}{c}{\hiveblue{Scratch}} & \multicolumn{3}{c}{\hiveblue{Pretrained}} & \multicolumn{3}{c}{\hiveblue{Scratch}}\\
 & \hiveblue{$v$} & \hiveblue{$x$} & \hiveblue{$x$--V} & \hiveblue{$v$} & \hiveblue{$x$} & \hiveblue{$x$--V} & \hiveblue{$v$} & \hiveblue{$x$} & \hiveblue{$x$--V} & \hiveblue{$v$} & \hiveblue{$x$} & \hiveblue{$x$--V}\\
\midrule
Adjust Bottle & 96 & 82 & 94 & 100 & 100 & 96 & 96 & 94 & 88 & 100 & 100 & 98\\
Beat Block Hammer & 92 & 66 & 64 & 96 & 90 & 80 & 92 & 82 & 60 & 96 & 86 & 84\\
Blocks Ranking RGB & 66 & 68 & 16 & 62 & 66 & 78 & 80 & 70 & 6 & 70 & 62 & 76\\
Blocks Ranking Size & 22 & 34 & 4 & 32 & 34 & 48 & 44 & 48 & 6 & 46 & 34 & 40\\
Click Alarmclock & 98 & 82 & 78 & 98 & 100 & 80 & 96 & 86 & 72 & 98 & 98 & 84\\
\bottomrule
\end{tabular}
\end{fitwidth}
\end{appendixresultstable}

\clearpage
\begin{appendixresultstable}{12}
\ContinuedFloat
\caption[]{FastWAM single-target NFE sweep, Clean success rates (\%): pretrained and scratch action heads, NFE=10, 16 (continued).}
\papertablelook
\begin{fitwidth}
\begin{tabular}{@{}T*{12}{N}@{}}
\toprule
\hiveblue{Task} & \multicolumn{6}{c}{\hiveblue{NFE=10}} & \multicolumn{6}{c}{\hiveblue{NFE=16}}\\
\cmidrule(lr){2-7} \cmidrule(lr){8-13}
 & \multicolumn{3}{c}{\hiveblue{Pretrained}} & \multicolumn{3}{c}{\hiveblue{Scratch}} & \multicolumn{3}{c}{\hiveblue{Pretrained}} & \multicolumn{3}{c}{\hiveblue{Scratch}}\\
 & \hiveblue{$v$} & \hiveblue{$x$} & \hiveblue{$x$--V} & \hiveblue{$v$} & \hiveblue{$x$} & \hiveblue{$x$--V} & \hiveblue{$v$} & \hiveblue{$x$} & \hiveblue{$x$--V} & \hiveblue{$v$} & \hiveblue{$x$} & \hiveblue{$x$--V}\\
\midrule
Adjust Bottle & 100 & 94 & 78 & 100 & 100 & 100 & 98 & 98 & 84 & 100 & 100 & 100\\
Beat Block Hammer & 92 & 78 & 56 & 94 & 92 & 76 & 96 & 88 & 38 & 98 & 90 & 90\\
Blocks Ranking RGB & 74 & 68 & 10 & 66 & 70 & 76 & 74 & 68 & 6 & 68 & 74 & 70\\
Blocks Ranking Size & 36 & 38 & 0 & 34 & 40 & 52 & 40 & 42 & 0 & 40 & 48 & 42\\
Click Alarmclock & 100 & 92 & 68 & 100 & 96 & 86 & 100 & 84 & 70 & 98 & 96 & 88\\
\bottomrule
\end{tabular}
\end{fitwidth}
\end{appendixresultstable}

\clearpage
\begin{appendixresultstable}{12}
\caption{FastWAM UniVEP NFE sweep, Clean success rates (\%): NFE=1, 2, 4. U=UniVEP, S=UniVEP scratch, X=UniVEP--X, V=UniVEP--X V-only.}\label{tab:app_fast_ufo}
\papertablelook
\begin{fitwidth}
\begin{tabular}{@{}T*{12}{N}@{}}
\toprule
\hiveblue{Task} & \multicolumn{4}{c}{\hiveblue{NFE=1}} & \multicolumn{4}{c}{\hiveblue{NFE=2}} & \multicolumn{4}{c}{\hiveblue{NFE=4}}\\
\cmidrule(lr){2-5} \cmidrule(lr){6-9} \cmidrule(lr){10-13}
 & \hiveblue{U} & \hiveblue{S} & \hiveblue{X} & \hiveblue{V} & \hiveblue{U} & \hiveblue{S} & \hiveblue{X} & \hiveblue{V} & \hiveblue{U} & \hiveblue{S} & \hiveblue{X} & \hiveblue{V}\\
\midrule
Adjust Bottle & 100 & 98 & 94 & 98 & 100 & 98 & 96 & 96 & 98 & 100 & 96 & 98\\
Beat Block Hammer & 88 & 78 & 86 & 84 & 92 & 84 & 90 & 86 & 92 & 64 & 92 & 88\\
Blocks Ranking RGB & 50 & 56 & 66 & 36 & 40 & 52 & 58 & 36 & 46 & 54 & 52 & 32\\
Blocks Ranking Size & 10 & 16 & 18 & 20 & 20 & 30 & 26 & 20 & 12 & 18 & 26 & 30\\
Click Alarmclock & 84 & 94 & 94 & 92 & 80 & 92 & 92 & 86 & 86 & 94 & 94 & 96\\
Click Bell & 98 & 68 & 74 & 88 & 98 & 76 & 78 & 84 & 88 & 70 & 80 & 80\\
\bottomrule
\end{tabular}
\end{fitwidth}
\end{appendixresultstable}

\clearpage
\begin{appendixresultstable}{12}
\ContinuedFloat
\caption[]{FastWAM UniVEP NFE sweep, Clean success rates (\%): NFE=8, 10, 16. U=UniVEP, S=UniVEP scratch, X=UniVEP--X, V=UniVEP--X V-only (continued).}
\papertablelook
\begin{fitwidth}
\begin{tabular}{@{}T*{12}{N}@{}}
\toprule
\hiveblue{Task} & \multicolumn{4}{c}{\hiveblue{NFE=8}} & \multicolumn{4}{c}{\hiveblue{NFE=10}} & \multicolumn{4}{c}{\hiveblue{NFE=16}}\\
\cmidrule(lr){2-5} \cmidrule(lr){6-9} \cmidrule(lr){10-13}
 & \hiveblue{U} & \hiveblue{S} & \hiveblue{X} & \hiveblue{V} & \hiveblue{U} & \hiveblue{S} & \hiveblue{X} & \hiveblue{V} & \hiveblue{U} & \hiveblue{S} & \hiveblue{X} & \hiveblue{V}\\
\midrule
Adjust Bottle & 100 & 96 & 96 & 98 & 100 & 98 & 98 & 96 & 100 & 100 & 98 & 98\\
Beat Block Hammer & 92 & 74 & 92 & 90 & 88 & 72 & 90 & 90 & 90 & 66 & 90 & 94\\
Blocks Ranking RGB & 48 & 50 & 56 & 46 & 40 & 48 & 48 & 50 & 42 & 50 & 50 & 44\\
Blocks Ranking Size & 16 & 18 & 22 & 26 & 28 & 18 & 26 & 24 & 14 & 14 & 28 & 20\\
Click Alarmclock & 74 & 90 & 94 & 90 & 78 & 94 & 96 & 96 & 76 & 86 & 98 & 82\\
Click Bell & 92 & 60 & 82 & 82 & 78 & 74 & 84 & 70 & 80 & 68 & 76 & 88\\
\bottomrule
\end{tabular}
\end{fitwidth}
\end{appendixresultstable}

\clearpage
\begin{appendixresultstable}{12}
\caption{LiLa Full UniVEP NFE sweep, Clean success rates (\%). U=UniVEP, X=UniVEP--X.}\label{tab:app_lila_ufo}
\papertablelook
\begin{fitwidth}
\begin{tabular}{@{}T*{12}{N}@{}}
\toprule
\hiveblue{Task} & \multicolumn{2}{c}{\hiveblue{NFE=1}} & \multicolumn{2}{c}{\hiveblue{NFE=2}} & \multicolumn{2}{c}{\hiveblue{NFE=4}} & \multicolumn{2}{c}{\hiveblue{NFE=8}} & \multicolumn{2}{c}{\hiveblue{NFE=10}} & \multicolumn{2}{c}{\hiveblue{NFE=16}}\\
\cmidrule(lr){2-3} \cmidrule(lr){4-5} \cmidrule(lr){6-7} \cmidrule(lr){8-9} \cmidrule(lr){10-11} \cmidrule(lr){12-13}
 & \hiveblue{U} & \hiveblue{X} & \hiveblue{U} & \hiveblue{X} & \hiveblue{U} & \hiveblue{X} & \hiveblue{U} & \hiveblue{X} & \hiveblue{U} & \hiveblue{X} & \hiveblue{U} & \hiveblue{X}\\
\midrule
Adjust Bottle & 100 & 100 & 100 & 100 & 100 & 100 & 100 & 100 & 100 & 100 & 100 & 100\\
Beat Block Hammer & 78 & 92 & 100 & 90 & 100 & 88 & 100 & 94 & 100 & 94 & 100 & 100\\
Blocks Ranking RGB & 98 & 98 & 96 & 92 & 100 & 100 & 100 & 96 & 100 & 96 & 100 & 96\\
Blocks Ranking Size & 92 & 88 & 86 & 84 & 78 & 82 & 86 & 78 & 86 & 78 & 82 & 82\\
Click Alarmclock & 46 & 38 & 72 & 62 & 86 & 70 & 76 & 76 & 88 & 74 & 86 & 78\\
Click Bell & 60 & 26 & 92 & 66 & 98 & 100 & 94 & 94 & 96 & 98 & 90 & 100\\
\bottomrule
\end{tabular}
\end{fitwidth}
\end{appendixresultstable}

\clearpage
\begin{appendixresultstable}{12}
\caption{LiLa Full UniVEP NFE sweep, Randomized success rates (\%). U=UniVEP, X=UniVEP--X.}\label{tab:app_lila_randomized}
\papertablelook
\begin{fitwidth}
\begin{tabular}{@{}T*{12}{N}@{}}
\toprule
\hiveblue{Task} & \multicolumn{2}{c}{\hiveblue{NFE=1}} & \multicolumn{2}{c}{\hiveblue{NFE=2}} & \multicolumn{2}{c}{\hiveblue{NFE=4}} & \multicolumn{2}{c}{\hiveblue{NFE=8}} & \multicolumn{2}{c}{\hiveblue{NFE=10}} & \multicolumn{2}{c}{\hiveblue{NFE=16}}\\
\cmidrule(lr){2-3} \cmidrule(lr){4-5} \cmidrule(lr){6-7} \cmidrule(lr){8-9} \cmidrule(lr){10-11} \cmidrule(lr){12-13}
 & \hiveblue{U} & \hiveblue{X} & \hiveblue{U} & \hiveblue{X} & \hiveblue{U} & \hiveblue{X} & \hiveblue{U} & \hiveblue{X} & \hiveblue{U} & \hiveblue{X} & \hiveblue{U} & \hiveblue{X}\\
\midrule
Adjust Bottle & 100 & 100 & 100 & 100 & 100 & 100 & 100 & 100 & 100 & 100 & 100 & 100\\
Beat Block Hammer & 76 & 98 & 86 & 88 & 90 & 88 & 94 & 92 & 86 & 88 & 96 & 92\\
Blocks Ranking RGB & 100 & 96 & 100 & 98 & 100 & 100 & 100 & 100 & 96 & 98 & 100 & 98\\
Blocks Ranking Size & 74 & 76 & 72 & 76 & 76 & 72 & 72 & 72 & 72 & 70 & 74 & 68\\
Click Alarmclock & 58 & 38 & 78 & 72 & 82 & 84 & 78 & 82 & 78 & 86 & 80 & 88\\
Click Bell & 62 & 40 & 84 & 66 & 92 & 86 & 92 & 92 & 94 & 92 & 90 & 92\\
\bottomrule
\end{tabular}
\end{fitwidth}
\end{appendixresultstable}

\clearpage
\begin{appendixresultstable}{12}
\setlength{\appresultnumberwidth}{27.5pt}
\caption{Fixed-checkpoint route controls on RoboTwin Clean: complete 50-task success rates (\%). StarVLA uses \method{}; all FastWAM controls use \method{}--X.}\label{tab:app_route_ablation}
\papertablelook
\begin{fitwidth}
\begin{tabular}{@{}T*{12}{N}@{}}
\toprule
\hiveblue{Task} & \multicolumn{2}{c}{\hiveblue{StarVLA}} & \multicolumn{10}{c}{\hiveblue{FastWAM}}\\
 & \multicolumn{2}{c}{\hiveblue{NFE=4}} & \multicolumn{3}{c}{\hiveblue{NFE=8}} & \multicolumn{4}{c}{\hiveblue{NFE=10}} & \multicolumn{3}{c}{\hiveblue{NFE=16}}\\
\cmidrule(lr){2-3} \cmidrule(lr){4-6} \cmidrule(lr){7-10} \cmidrule(lr){11-13}
 & \hiveblue{\mbox{$X\!\to\!V$}} & \hiveblue{\mbox{$V\!\to\!X$}} & \hiveblue{\mbox{$X\!\to\!V$}} & \hiveblue{\mbox{ X-only}} & \hiveblue{\mbox{ V-only}} & \hiveblue{\mbox{$X\!\to\!V$}} & \hiveblue{\mbox{$V\!\to\!X$}} & \hiveblue{\mbox{ X-only}} & \hiveblue{\mbox{ V-only}} & \hiveblue{\mbox{$X\!\to\!V$}} & \hiveblue{\mbox{ X-only}} & \hiveblue{\mbox{ V-only}}\\
\midrule
Adjust Bottle & 100 & 100 & 96 & 98 & 98 & 98 & 98 & 98 & 96 & 98 & 96 & 98\\
Beat Block Hammer & 94 & 96 & 92 & 86 & 90 & 90 & 78 & 84 & 90 & 90 & 82 & 94\\
Blocks Ranking RGB & 70 & 78 & 56 & 46 & 46 & 48 & 54 & 52 & 50 & 50 & 44 & 44\\
Blocks Ranking Size & 30 & 42 & 22 & 22 & 26 & 26 & 22 & 20 & 24 & 28 & 20 & 20\\
Click Alarmclock & 0 & 0 & 94 & 84 & 90 & 96 & 96 & 92 & 96 & 98 & 98 & 82\\
\bottomrule
\end{tabular}
\end{fitwidth}
\end{appendixresultstable}

\clearpage
\subsection{LiLa-WAM single-target sweeps}
\label{sec:app_lila_single_nfe}
\begin{table}[!htbp]
\centering
\begin{minipage}{.84\linewidth}
\caption{LiLa-WAM, single-target V. Per-task RoboTwin Clean success (\%) at each call budget.}
\label{tab:app_lila_single_nfe_v}
\papertablestyle\fontsize{9}{10.5}\selectfont
\begin{tabularx}{\linewidth}{@{}PRRRRRR@{}}
\toprule
\hiveblue{Task} & \hiveblue{1} & \hiveblue{2} & \hiveblue{4} & \hiveblue{8} & \hiveblue{10} & \hiveblue{16}\\
\midrule
Adjust Bottle & 100 & 100 & 100 & 100 & 100 & 100\\
Beat Block Hammer & 80 & 90 & 80 & 82 & 92 & 84\\
Blocks Ranking RGB & 88 & 96 & 92 & 90 & 96 & 96\\
Blocks Ranking Size & 56 & 76 & 66 & 70 & 80 & 72\\
Click Alarmclock & 48 & 76 & 78 & 84 & 84 & 80\\
Click Bell & 60 & 68 & 88 & 90 & 90 & 84\\
Dump Bin Bigbin & 88 & 92 & 90 & 86 & 90 & 92\\
\bottomrule
\end{tabularx}
\end{minipage}
\end{table}
\clearpage
\begin{table}[!htbp]
\centering
\begin{minipage}{.84\linewidth}
\caption{LiLa-WAM, single-target X. Per-task RoboTwin Clean success (\%) at each call budget.}
\label{tab:app_lila_single_nfe_x}
\papertablestyle\fontsize{9}{10.5}\selectfont
\begin{tabularx}{\linewidth}{@{}PRRRRRR@{}}
\toprule
\hiveblue{Task} & \hiveblue{1} & \hiveblue{2} & \hiveblue{4} & \hiveblue{8} & \hiveblue{10} & \hiveblue{16}\\
\midrule
Adjust Bottle & 100 & 100 & 100 & 100 & 100 & 100\\
Beat Block Hammer & 92 & 90 & 78 & 88 & 94 & 84\\
Blocks Ranking RGB & 94 & 96 & 94 & 92 & 92 & 96\\
Blocks Ranking Size & 76 & 72 & 78 & 80 & 88 & 82\\
Click Alarmclock & 44 & 50 & 48 & 44 & 50 & 58\\
Click Bell & 38 & 22 & 32 & 38 & 80 & 96\\
Dump Bin Bigbin & 88 & 92 & 88 & 90 & 92 & 88\\
\bottomrule
\end{tabularx}
\end{minipage}
\end{table}
\clearpage
\begin{table}[!htbp]
\centering
\begin{minipage}{.84\linewidth}
\caption{LiLa-WAM, single-target X--V. Per-task RoboTwin Clean success (\%) at each call budget.}
\label{tab:app_lila_single_nfe_xV}
\papertablestyle\fontsize{9}{10.5}\selectfont
\begin{tabularx}{\linewidth}{@{}PRRRRRR@{}}
\toprule
\hiveblue{Task} & \hiveblue{1} & \hiveblue{2} & \hiveblue{4} & \hiveblue{8} & \hiveblue{10} & \hiveblue{16}\\
\midrule
Adjust Bottle & 100 & 100 & 100 & 100 & 100 & 100\\
Beat Block Hammer & 92 & 94 & 92 & 90 & 92 & 88\\
Blocks Ranking RGB & 96 & 94 & 96 & 96 & 96 & 94\\
Blocks Ranking Size & 78 & 82 & 80 & 80 & 78 & 86\\
Click Alarmclock & 60 & 84 & 82 & 80 & 78 & 80\\
Click Bell & 74 & 84 & 92 & 92 & 88 & 90\\
Dump Bin Bigbin & 90 & 82 & 94 & 92 & 94 & 90\\
\bottomrule
\end{tabularx}
\end{minipage}
\end{table}
\clearpage

\clearpage
\subsection{LiLa-WAM with the sampling order reversed}
\label{sec:app_lila_reverse}

\begin{table}[!htbp]
\centering
\begin{minipage}{.84\linewidth}
\caption{Per-task RoboTwin Clean success (\%) for the fixed LiLa-WAM UniVEP checkpoint under $V\to X$ sampling at 10 NFE.}
\label{tab:app_lila_reverse}
\papertablestyle\footnotesize
\begin{tabularx}{\linewidth}{@{}PRPR@{}}
\toprule
\hiveblue{Task} & \hiveblue{$\boldsymbol{V\!\to\!X}$} & \hiveblue{Task} & \hiveblue{$\boldsymbol{V\!\to\!X}$}\\
\midrule
Adjust Bottle & 100 & Place Can Basket & 86\\
Beat Block Hammer & 98 & Place Cans Plasticbox & 96\\
Blocks Ranking RGB & 100 & Place Container Plate & 68\\
Blocks Ranking Size & 82 & Place Dual Shoes & 44\\
Click Alarmclock & 82 & Place Empty Cup & 100\\
Click Bell & 88 & Place Fan & 90\\
Dump Bin Bigbin & 98 & Place Mouse Pad & 80\\
Grab Roller & 100 & Place Object Basket & 82\\
Handover Block & 90 & Place Object Scale & 84\\
Handover Mic & 86 & Place Object Stand & 96\\
Hanging Mug & 54 & Place Phone Stand & 96\\
Lift Pot & 100 & Place Shoe & 92\\
Move Can Pot & 90 & Press Stapler & 100\\
Move Pillbottle Pad & 92 & Put Bottles Dustbin & 92\\
Move Playingcard Away & 100 & Put Object Cabinet & 68\\
Move Stapler Pad & 70 & Rotate QRcode & 90\\
Open Laptop & 92 & Scan Object & 86\\
Open Microwave & 44 & Shake Bottle Horizontally & 100\\
Pick Diverse Bottles & 80 & Shake Bottle & 100\\
Pick Dual Bottles & 98 & Stack Blocks Three & 84\\
Place A2B Left & 58 & Stack Blocks Two & 98\\
Place A2B Right & 64 & Stack Bowls Three & 82\\
Place Bread Basket & 76 & Stack Bowls Two & 96\\
Place Bread Skillet & 70 & Stamp Seal & 42\\
Place Burger Fries & 82 & Turn Switch & 70\\
\midrule
\multicolumn{3}{@{}P}{\textbf{Mean over all 50 tasks}} & \textbf{84.32}\\
\bottomrule
\end{tabularx}
\end{minipage}
\end{table}

\end{document}